\pdfoutput=1

\documentclass[11pt]{article}

\usepackage[preprint]{acl}

\usepackage{times}
\usepackage{latexsym}
\usepackage[table]{xcolor}
\definecolor{band}{RGB}{246,246,246} 
\usepackage[T1]{fontenc}

\usepackage[utf8]{inputenc}

\usepackage{microtype}

\usepackage{inconsolata}

\usepackage{graphicx}

\usepackage[utf8]{inputenc} 
\usepackage[T1]{fontenc}    

\usepackage{booktabs}       
\usepackage{amsfonts}       
\usepackage{nicefrac}       
\usepackage{microtype}      
\usepackage{multirow}
\usepackage{makecell}
\usepackage[most]{tcolorbox}
\usepackage{xcolor}
\usepackage{amsmath}
\usepackage{amsthm}
\usepackage{enumitem}
\theoremstyle{remark}

\usepackage{listings}
\tcbuselibrary{breakable}
\usepackage{pifont}
\newcommand{\xmark}{\ding{55}}  
\newtcolorbox{helpfulbox}{
  breakable,
  enhanced jigsaw,
  colback=green!5!white,
  colframe=green!50!black,
  arc=3mm,
  boxrule=0.8pt,
  title=Helpful Hints,
  before skip=6pt,
  after skip=6pt,
}

\newtcolorbox{conflictbox}{
  breakable,
  enhanced jigsaw,
  colback=yellow!10!white,
  colframe=yellow!80!black,
  arc=3mm,
  boxrule=0.8pt,
  title=Conflicting Hints,
  before skip=6pt,
  after skip=6pt,
}

\newtcolorbox{adversarialbox}{
  breakable,
  enhanced jigsaw,
  colback=red!5!white,
  colframe=red!50!black,
  arc=3mm,
  boxrule=0.8pt,
  title=Adversarial Hints,
  before skip=6pt,
  after skip=6pt,
}

\title{AssistedDS: Benchmarking How External Domain Knowledge Assists LLMs in Automated Data Science}

\author{%
 An Luo\textsuperscript{1\tiny*},
 Xun Xian\textsuperscript{1\tiny*},
 Jin Du\textsuperscript{1},
  Fangqiao Tian\textsuperscript{1},
  Ganghua Wang\textsuperscript{4},\\
  \bfseries
  Ming 
  Zhong\textsuperscript{5},
  Shengchun
  Zhao\textsuperscript{6},
  Xuan Bi\textsuperscript{1},
Zirui Liu\textsuperscript{1},
  Jiawei Zhou\textsuperscript{3},\\
  \bfseries
  Jayanth Srinivasa\textsuperscript{2},
  Ashish Kundu\textsuperscript{2},
  Charles Fleming\textsuperscript{2},
  Mingyi Hong\textsuperscript{1},
  Jie Ding\textsuperscript{1}
  \\
  \textsuperscript{1}University of Minnesota,
  \textsuperscript{2}Cisco Research,
  \textsuperscript{3}Stony Brook University\\
  \textsuperscript{4}University of Chicago,
  \textsuperscript{5}Independent Researcher,
  \textsuperscript{6}University of Michigan 
}

\begin{document}

\maketitle

\begingroup
  \renewcommand\thefootnote{\fnsymbol{footnote}}%
  \footnotetext[1]{Equal distribution.}%
\endgroup

\begin{abstract}
Large language models (LLMs) have advanced the automation of data science workflows. Yet it remains unclear whether they can critically leverage external domain knowledge as human data scientists do in practice.
To answer this question, we introduce AssistedDS (Assisted Data Science), a benchmark designed to systematically evaluate how LLMs handle domain knowledge in tabular prediction tasks. AssistedDS features both synthetic datasets with explicitly known generative mechanisms and real-world Kaggle competitions, each accompanied by curated bundles of helpful and adversarial documents. These documents provide domain-specific insights into data cleaning, feature engineering, and model selection.
We assess state-of-the-art LLMs on their ability to discern and apply beneficial versus harmful domain knowledge, evaluating submission validity, information recall, and predictive performance. Our results demonstrate three key findings: (1) LLMs frequently exhibit an uncritical adoption of provided information, significantly impairing their predictive performance when adversarial content is introduced, (2) helpful guidance is often insufficient to counteract the negative influence of adversarial information, and (3) in Kaggle datasets, LLMs often make errors in handling time-series data, applying consistent feature engineering across different folds, and interpreting categorical variables correctly.
These findings highlight a substantial gap in current models' ability to critically evaluate and leverage expert knowledge, underscoring an essential research direction for developing more robust, knowledge-aware automated data science systems. Our data and code are publicly available \href{https://github.com/jeremyxianx/Assisted-DS}{\textcolor{blue}{here}}. 
\end{abstract}

\section{Introduction}

Recent advancements in large language models (LLMs) have led to significant progress in automating data science workflows. LLMs such as GPT-4~\cite{achiam2023gpt} and Claude~\cite{anthropic2025claude3.7} have shown remarkable capabilities in generating code and performing machine learning tasks, enabling end-to-end automation of many routine data analysis procedures~\cite{Grosnit2024LargeLM, Hong2024DataIA,Jiang2025AIDEAE,Liang2025IMCTSEA}. 

Despite these capabilities, a critical aspect of LLMs for real-world data science remains underexplored: the effective integration and critical evaluation of external domain knowledge. In practice, data scientists do not simply run standard algorithms--they routinely incorporate domain-specific knowledge, solicit feedback from colleagues, and weigh multiple sources of information before finalizing decisions~\cite{Mao2019HowDS, Zhang2020HowDD}. Such processes are essential for handling the complexities and ambiguities inherent in real-world data tasks.

However, current research on LLM-driven data science has largely focused on code generation and pipeline execution~\cite{Li2024AutoKaggleAM, Jiang2025AIDEAE}, neglecting the crucial role of external expert knowledge in practical applications. Existing evaluation benchmarks~\cite{chan2025mlebench, jing2025dsbench, Zhang2025DataSciBenchAL} rarely measure how LLMs process, filter, or critically adopt domain knowledge, especially when such information may be misleading or adversarial. This raises fundamental questions: Can LLMs distinguish between helpful and harmful external domain knowledge in data science workflows? Or do they simply adopt all information, risking degraded performance when exposed to adversarial guidance?

\begin{figure}[htbp]
\centering
\includegraphics[width=0.45\textwidth]{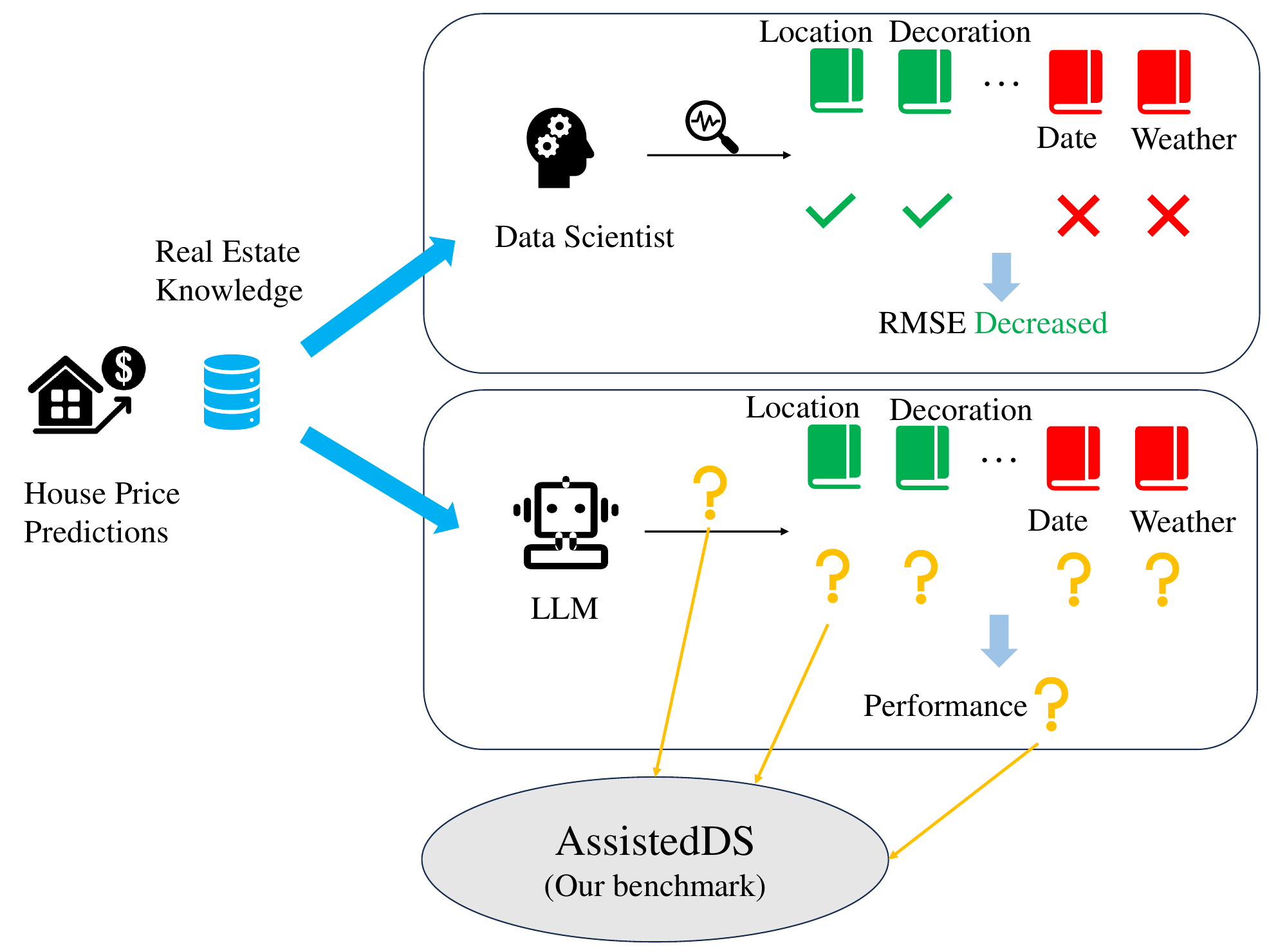}
\caption{How AssistedDS establishes its unique position. \textbf{Top}: A human data scientist analyses data, and often critically filters external domain knowledge. This means accepting helpful hints (green) and rejecting adversarial ones (red), which leads to demonstrably better results.
\textbf{Bottom}: An LLM can automate the same analytical steps, but its ability to judge the quality of domain knowledge is unknown.
AssistedDS evaluates whether an LLM, like its human counterpart, can identify and leverage helpful information while resisting harmful advice in automated data science workflow.}
\label{fig:abstract}
\end{figure}

In this paper, we introduce \textbf{AssistedDS} (\underline{A}ssisted \underline{D}ata \underline{S}cience), a benchmark designed specifically to measure the capability of LLMs to leverage domain knowledge effectively and critically for data science. AssistedDS evaluates LLM performance on tabular prediction tasks augmented with diverse domain knowledge bundles, which include both helpful domain insights and deliberately adversarial information. Each task within AssistedDS comes with synthetic datasets that have ground-truth generative mechanisms, enabling precise characterization of helpful versus adversarial domain knowledge. We complement these synthetic experiments with real-world Kaggle datasets~\cite{allstate-claims-severity, bike-sharing-demand, bnp-paribas-cardif-claims-management, otto-group-product-classification-challenge, rossmann-store-sales}, where helpfulness is defined by Kaggle-hosted highly-rated expert notebooks and adversarial examples are constructed from low-quality notebooks.

Our experiments reveal three critical findings. First, LLMs consistently demonstrate an uncritical tendency to follow provided information, even in the presence of adversarial cues. Our experiments on synthetic datasets show that even GPT-o4-mini fails to filter out $60\%$ of adversarial hints, and, in the worst case, GPT-4o-mini blindly follows every adversarial hints. All these blind following of adversarial hints leads to substantial performance declines, as much as $-159.41\%$. Second, prompts indicating helpful versus adversarial content provides only limited improvement, highlighting a fundamental limitation in current models' ability to critically evaluate external sources. Third, we find that LLM's use of domain knowledge for real-world data such as Kaggle has new issues like deficiencies in handling time-series data, leading to enormous failures code execution. For instance, in the Bike Sharing competition, LLMs generated 131 ``not found in axis'' failures ($68.95\%$ of all errors) when asked to predict outcomes from 20th to the end of the month from first 19 days of each month, with 62 of those errors ($47.33 \%$) explicitly mentioning \texttt{datetime}. This clearly shows that the models treat \texttt{datetime} as isolated categories instead of recognizing them as points on a continuous timeline.

These results underscore a crucial limitation in current LLM-driven automation: the lack of critical reasoning about external domain knowledge. Recognizing this shortcoming, we argue that future research must focus on developing LLM capabilities for critical assessment and selective integration of external domain knowledge. By highlighting these persistent weaknesses, AssistedDS lays foundational groundwork for robust, knowledge-aware automated data science.

Our main contributions are summarized below.
\begin{itemize}
    \item We propose \textbf{AssistedDS}, the first benchmark to evaluate LLMs’ ability to critically integrate and filter external domain knowledge in automated data science workflows, including both synthetic and real-world datasets.
    \item We curate controlled bundles of helpful and adversarial domain knowledge, enabling assessment of LLM's robustness under mixed and ambiguous informational environments.
    \item We present empirical evidence in both synthetic and real-world datasets that state-of-the-art LLMs frequently adopt external knowledge uncritically, suffering substantial performance degradation in the presence of adversarial information.
    \item We found that in real-world datasets LLMs often fail in time-series handling, feature engineering consistency, and categorical data interpretation, underscoring the need for knowledge-aware automation.
\end{itemize}

\subsection{Related Work}

\textbf{Automated Data Science with LLMs. }
Automating data science workflows using LLMs has seen considerable interest in recent research. AutoKaggle~\cite{Li2024AutoKaggleAM} structures the data science process into clearly defined stages, employing specialized agents for iterative coding and debugging. Agent K~\cite{Grosnit2024LargeLM} frames data science tasks as a Markov Decision Process, enabling agent-based interactions that include both internal reasoning and external actions. Similarly, Data Interpreter~\cite{Hong2024DataIA} models the data science workflow dynamically as a hierarchical graph, optimizing performance by iteratively refining task dependencies. SPIO~\cite{Seo2025SPIOEA} uses a central planner to explore multiple predictive strategies, either selecting a single best strategy or forming ensembles to achieve robust performance. Despite these advances, current automated workflows rarely evaluate the integration and impact of external domain knowledge.

\textbf{Evaluation Benchmarks. }
Benchmarks like MLE-bench~\cite{chan2025mlebench} focus on evaluating end-to-end predictive performance of machine learning pipelines. DA-Bench ~\cite{Hu2024InfiAgentDABenchEA} and DA-code~\cite{huang-etal-2024-da} benchmark code generation in data science. DSBench~\cite{jing2025dsbench} and DataSciBench~\cite{Zhang2025DataSciBenchAL} expand these evaluations beyond mere predictive accuracy, probing reasoning capabilities through curated tasks that involve interpretation and logical assessment of data. However, these benchmarks do not explicitly examine how well LLMs incorporate external guidance, particularly when such information varies in quality or reliability. As shown in Table~\ref{tab:bench-compare}, AssistedDS is the only benchmark to include both helpful and adversarial external knowledge across synthetic and real-world datasets, which provides a unique lens on LLMs' ability to critically leverage domain knowledge on data science.


\begin{table*}[t]
\centering
\caption{Among related benchmarks, AssistedDS uniquely evaluates critical use of external guidance on end-to-end code generation for data science with both synthetic and real-world data.}
\label{tab:bench-compare}
\scalebox{0.75}{
\begin{tabular}{lcccc}
\toprule
\textbf{Benchmark} & \textbf{External Knowledge}  & \textbf{End-to-End Code} &  \textbf{Synthetic Data} &  \textbf{Real-World Data}  \\
\midrule
\textbf{AssistedDS (Ours)} & \ding{51} & \ding{51} & \ding{51} & \ding{51} \\
MLE\,-bench~\cite{chan2025mlebench} & \xmark & \ding{51} & \ding{51} & \ding{51}\\
DA\,-Bench~\cite{Hu2024InfiAgentDABenchEA} & \xmark & \ding{51} & \ding{51} & \ding{51} \\
DA\,-Code~\cite{huang-etal-2024-da} & \xmark & \ding{51} & \xmark & \ding{51} \\
DSBench~\cite{jing2025dsbench} & \xmark & \ding{51} & \xmark & \ding{51} \\
DataSciBench~\cite{Zhang2025DataSciBenchAL} & \xmark & \ding{51} & \xmark & \ding{51} \\
\bottomrule
\end{tabular}
}
\end{table*}

\textbf{Reasoning and Critical Evaluation in LLMs. }
Recent studies addressing reasoning capabilities in LLM-driven workflows emphasize iterative tree-based search methods. AIDE~\cite{Jiang2025AIDEAE}, SELA~\cite{Chi2024SELATE}, and I-MCTS~\cite{Liang2025IMCTSEA} leverage structured search strategies to iteratively refine solutions, suggesting improved reasoning about complex tasks. DS-Agent~\cite{guo2024dsagent} retrieves high-quality external knowledge sources, such as Kaggle notebooks from experts, to enable case-based reasoning. Yet, these approaches primarily demonstrate positive use cases and lack systematic analysis of LLMs' vulnerability to adversarial or misleading external information.

\textbf{Retrieval-Augmented Generation and Long Context Models.}
To reduce hallucination and improve factual accuracy, recent research has employed retrieval-augmented generation (RAG)~\cite{lewis2020retrieval,khandelwal2019generalization} and long context (LC) modeling\cite{chen2023extending, wang2024beyond}. LC methods generally outperform RAG for structured, dense contexts, effectively handling well-defined information such as notebook tutorials~\cite{li2024long}. Given the structured nature of external data science knowledge, our study adopts the LC approach by directly incorporating comprehensive external context into LLM prompts.
Our work addresses crucial gaps in the existing literature by benchmarking the capability of LLMs to integrate external domain expertise critically.

Our work contributes by examining the critical capacity of LLMs to leverage external domain knowledge for automated data science, evaluating their performance against both beneficial and adversarial information. This focused evaluation sets our work apart from existing literature, highlighting both the strengths and fundamental limitations of current automated data science approaches.

\section{Benchmark Curation}
\begin{figure*}[htbp]
\centering
\includegraphics[width=1.0\textwidth]{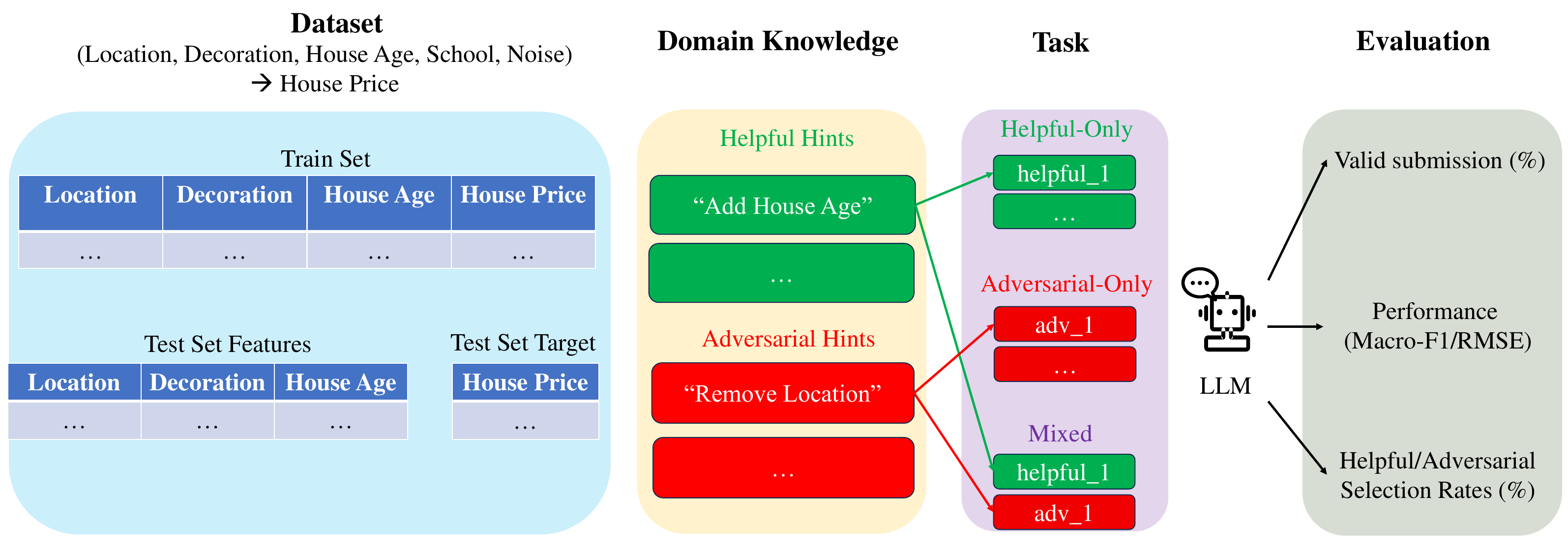}
\caption{
Overview of the AssistedDS Benchmark. \textbf{Left: Dataset.} We use datasets (synthetic or real-world) with defined features (e.g., Location, Decoration, House Age, School Quality) and clear train/test splits for tabular prediction tasks. \textbf{Center-left: Domain Knowledge.} For each dataset, we curate bundles of domain knowledge: helpful hints (e.g., “Add School Quality”) and adversarial hints (e.g., “Remove Location”), reflecting realistic expert guidance and potential pitfalls. \textbf{Center: Task Bundles.} Hints are grouped into tasks: Helpful-Only, Adversarial-Only, Mixed, and so on. Each represents different real-world information environments. \textbf{Center-right: LLM Execution.} An LLM receives both the data and bundled hints, together with the prompt, and generates a end-to-end code script. \textbf{Right: Evaluation.} We assess each model’s generated code by (i) valid submission rate, (ii) predictive performance (Macro-F1, RMSE, etc), and (iii) the rates at which helpful or adversarial hints are actually adopted in the generated code.}
\label{fig:overarching}
\end{figure*}

Our benchmark, \textbf{AssistedDS},  evaluates how effectively and critically LLMs leverage domain knowledge in automated data science workflows. An overview of our benchmark framework is illustrated in Figure \ref{fig:overarching}.  
We consider synthetic datasets to have control on domain knowledge, and Kaggle datasets to reflect real-world complexities on domain knowledge.  For synthetic datasets we generate dataset with transparent generating mechanism, and then where we can define and inject ``helpful'' hints that boost performance and ``adversarial'' hints that can cause performance decline. For Kaggle datasets we draw on real-world problems from Kaggle competitions, pairing each dataset with high-quality community notebooks that embody genuine domain knowledge through feature engineering, data cleaning, and analysis. To mirror realistic scenario for LLM's access to domain knowledge, we then designed tasks, which are different bundles of helpful and adversarial domain knowledge, paired with certain prompt template to let LLM generate an end-to-end code for submission.csv. Then we evaluate based on the code script generated by the LLM. Below, we elaborate more on why and how we prepare the datasets, domain knowledge, and evaluation metrics, and in Section \ref{sec:tasks} we describe the tasks as experimental settings.

\subsection{Datasets}\label{section:datasets}

Our benchmark encompasses both synthetic and real-world datasets. The synthetic datasets give us full control over the domain knowledge: we construct feature–label relationships to produce helpful hints (which should improve predictive performance) and adversarial hints (which should degrade them). The real-world collection consists of competitions drawn from Kaggle, each accompanied by community notebooks rated by us that capture authentic domain expertise--ranging from data preprocessing to feature manipulation. In the sections that follow, we describe how synthetic datasets and Kaggle datasets are curated.

\paragraph{Synthetic Datasets.} We curate 10 synthetic tabular datasets, encompassing 6 classification and 4 regression tasks. Each dataset is generated with a clearly defined data-generating process involving realistic domain-specific features, nonlinear relationships, interactions, noise, and intentional data imperfections such as missing values and outliers. The datasets cover diverse application domains including Game Revenue, Real Estate, Second-hand Goods, Power Generation, Diabetes, Machine Failure, Haircut Rate, Wine Quality, Song Popularity, and Housekeeping. For each dataset, we provide standard training/test splits, along with additional files for new or manipulated features for domain knowledge preparation. A summary of synthetic datasets we curated is in Appendix \ref{appendix:datasets}.

\textbf{Illustrative Example: Song Popularity}
We illustrate our synthetic dataset generation using the ``Song Popularity'' domain. Domain-relevant features are first specified, followed by the response variable (\emph{Popularity Class}) capturing realistic relationships and stochasticity. Specifically, the \emph{Popularity Score} is generated from a linear combination of domain-specific features such as Artist Fame (positive effect), Genre (pop and hip-hop positively affect popularity, other genres negatively), Danceability (strong quadratic positive effect), Tempo in BPM (positive), high Social Media Hype (positive binary effect), presence of Featuring Artist interacting positively with Artist Fame, squared Lyric Sentiment (negative quadratic effect), Music Video Budget (positive), and Album Position (negative). A logistic noise term is added to mimic real-world stochasticity. This continuous score is then discretized into three classes: \emph{Unpopular}, \emph{Ordinary}, and \emph{Hot}, based on predetermined thresholds. The details of how we generated this dataset is in Appendix \ref{appendix:dataset_generation}. We then hold out two features, Music Video Budget and Tempo BPM, for domain knowledge creation, as explained in Section \ref{sec:domainsyn}.

\paragraph{Kaggle Datasets.} To capture real-world challenges in both datasets and domain knowledge that synthetic datasets could not exhibit, we include Kaggle datasets as a complement to synthetic datasets in our benchmark. Importantly, we do not simply replicate Kaggle competitions. Instead, we carefully subsample the released datasets to reduce overlap with tasks that may have appeared in LLMs’ pretraining, thereby mitigating the risk of overfitting due to memorization. The detailed procedures are provided in Appendix~\ref{appendix:advnotebook}. Beyond data, we also extract and curate community-authored Kaggle notebooks as a source of domain knowledge, which we systematically rate and rank (see Section~\ref{sec:kaggleknowledge}) to support downstream evaluation. We select five widely participated competitions: Allstate Claims Severity~\cite{allstate-claims-severity}, Bike Sharing Demand~\cite{bike-sharing-demand}, BNP Paribas Cardif Claims Management~\cite{bnp-paribas-cardif-claims-management}, Otto Group Product Classification Challenge~\cite{otto-group-product-classification-challenge}, and Rossmann Store Sales~\cite{rossmann-store-sales}. For each dataset, we construct a fixed train/test split and release three files: \texttt{train.csv}, \texttt{test.csv}, and a \texttt{test\_target.csv} containing identifiers and ground truth labels used exclusively for evaluation. Rossmann also includes a file \texttt{store.csv}, as in the original Kaggle competition.

Detailed dataset preparation procedures and examples are provided in Appendix~\ref{appendix:dataset_generation} and \ref{appendix:advnotebook}.

\subsection{Domain Knowledge (Hints)}\label{sec:domain knowledge}

Domain knowledge plays a central role in our benchmark, as it enables us to evaluate how LLMs integrate external information in automated data science workflows. While we introduced the use of domain knowledge in Section~\ref{section:datasets}, we now provide further details on its construction and use.

To assess whether LLMs can critically utilize domain knowledge when generating end-to-end code for tabular prediction tasks, we curate two distinct types of domain knowledge: helpful and adversarial. An LLM that critically engages with domain knowledge should selectively incorporate helpful hints while rejecting misleading ones, leading to improved prediction performance. In contrast, an LLM that uses domain knowledge uncritically--absorbing any hint indiscriminately--is likely to suffer degraded performance.

Below, we detail the procedures for preparing domain knowledge in synthetic and Kaggle datasets.

\subsubsection{Domain knowledge:  Synthetic Datasets}\label{sec:domainsyn}

For synthetic datasets, helpful hints truthfully reflect data-generating processes or proven domain knowledge, while adversarial hints intentionally mislead by advocating inappropriate modeling choices. Hints are combined into bundles with controlled compositions to simulate various realistic informational environments. 

Take Song Popularity dataset as an example again.  Based the  generating mechanism, we provide specific hints aimed at helpful or adversarial modeling choices:

\begin{helpfulbox}
\textbf{Helpful Hints} encourage the addition of predictive features previously withheld:
\begin{itemize}[left=0pt,label=\(\bullet\)]
  \item Add feature \texttt{Music-Video-Budget}: Estimated music video production budget (units of 10,000 dollars, continuous, [0, 500]).
  \item Add feature \texttt{Tempo-BPM}: Tempo of the song (continuous, [60, 200] BPM).
\end{itemize}
\end{helpfulbox}

\begin{adversarialbox}
\textbf{Adversarial Hints} mislead by suggesting removal of important predictive features:
\begin{itemize}[left=0pt,label=\(\bullet\)]
  \item Remove (crucial predictive) feature \texttt{in-game-purchases}.
  \item Remove (crucial predictive) feature \texttt{monetization-model-Premium}.
\end{itemize}
\end{adversarialbox}

\subsubsection{Domain Knowledge: Kaggle Datasets}\label{sec:kaggleknowledge}

To reflect real-world domain knowledge, we use the community notebooks~\cite{KaggleNotebooksDocs2025} from Kaggle platform as domain knowledge. These notebooks often give insights like how features should be created to boost performance. 
To collect high-quality notebooks, we collected the \textit{Code} section of each competition and ranked available notebooks by vote count. We then retrieved the top 50 notebooks for each competition. For each notebook, we extracted source code, number of votes, number of comments and model performance (e.g., MAE, accuracy), and then designed a composite quality score to evaluate the informativeness and structure of each notebook. 
The score for each notebook is calculated as
\begin{equation*}
0.3 \times \mathcal{S} + 0.5 \times \mathcal{V} + 0.2 \times \mathcal{C},
\end{equation*}
where $\mathcal{S}$ is the normalized log-performance, $\mathcal{V}$ is the normalized log-vote count, and $\mathcal{C}$ is the normalized log-comment count for each notebook.
Normalization was performed using min-max scaling across all collected notebooks. Based on these scores, we selected the top 4 notebooks as high-quality exemplars and set them as helpful domain knowledge. For adversarial domain knowledge, we take the bottom 4 as low-quality notebooks for each competition, and remove crucial feature in the code to make them adversarial. See details in Appendix~\ref{appendix:advnotebook}.

\subsection{Protocols and Models Used for Evaluation}

With the datasets and corresponding domain knowledge in place, we next assess the LLMs’ end-to-end automated data science capabilities. Specifically, we evaluate the Python code generated from a single prompt. Each prompt includes a dataset description, a dataset preview, and a bundle of domain knowledge (as specified in Section \ref{sec:experimenttask}).
In this paper, we evaluate three LLMs from OpenAI: GPT-o4-mini, a cost-effective model that delivers reasoning capabilities, GPT-4o, the flagship model, and GPT-4o-mini, a more compact variant of GPT-4o. In addition to these three LLMs from OpenAI, we also evalute Claude 3.5 haiku, Gemini 2.0 flash, and DeepSeek-chat. Meanwhile, we include a human baseline, where a human data scientist critically selects from the provided domain knowledge using cross-validation.

\subsection{Evaluation Metrics}
\label{sec:metrics}
\begin{itemize}[leftmargin=*]
    \item \textbf{Valid Submission} (\%): Proportion of runs where \texttt{submission.csv} is produced and formatted correctly for evaluation with \texttt{test\_target.csv}. This is to check code-execution.
    \item \textbf{Helpful Selection} (\%): Number of helpful hints used in the generated code $/$ number of helpful hints provided.  This is to measure how helpful domain knowledge is used by LLMs under different settings, and the higher the better. 
        \item \textbf{Adversarial Selection} (\%): Number of adversarial hints used in the generated code $/$ number of adversarial hint provided. This is to measure how adversarial domain knowledge is used by LLMs under different settings, and the lower the better.
    \item \textbf{Performance}: For synthetic datasets, we use RMSE for regression and macro-F1 for classification.  This is to see how predictive performance would change when different bundles of domain knowledge are provided. We only report this for synthetic datasets. To indicate the change, we also report \textbf{Performance Change}, which is the percentile change from the case where no domain knowledge is provided. We adjust it for RMSE so that for both regression and classification positive values indicate improvement and negative values indicate degradation. For Kaggle datasets, we use the performance metrics defined by each competition, as detailed in Appendix~\ref{appendix:kaggleresults}, and the performance change is adjusted similarly.
\end{itemize}

For both helpful and adversarial selections, we employ GPT-4.1 model as an evaluator to assess which hints are actually understood and incorporated in the code generated by the LLM. We confirmed its agreement with graduate-level human judgments in Appendix~\ref{appendix:judge}.

\section{Experimental Studies}\label{sec:experimenttask}

To test an LLM’s ability to critically leverage both helpful and adversarial domain knowledge curated as described in Section \ref{sec:domain knowledge}, we design experiments (tasks) using various combinations of helpful and adversarial hints together with different prompt configurations. The overall design principle is to vary both the proportion of helpful versus adversarial hints and the degree to which prompts inform the LLM about the quality of these hints, allowing us to evaluate how LLMs distinguish and utilize domain knowledge under realistic conditions.

\subsection{Tasks}
\label{sec:tasks}

For synthetic datasets, we design five tasks:
\begin{itemize}
    \item \textbf{None } This task feeds no domain knowledge to LLMs. Such case would indicate how LLMs perform when no domain knowledge is present for decision-making, which would serve as a baseline performance.
    \item \textbf{Helpful‑Only }
    consists of $n\in\{1, 2\}$ helpful hints only. We use \textbf{neutral prompt} here, i.e., we do not provide quality information regarding domain knowledge.  This task is to investigate how well LLMs can follow helpful domain knowledge when this is the only source. Ideally this would ensure the best performance if LLMs takes all helpful advice. 
    \item \textbf{Adversarial-Only} consists of $n\in\{1, 2\}$ adversarial hints only. We use \textbf{neutral prompt} here, i.e., we do not provide quality information regarding domain knowledge. If LLMs just blindly follow adversarial hints in their code, their predictive performance would drop by our design of ``adversarial'' hint. 
    \item \textbf{Mixed } consists of mixed helpful and adversarial hints, with count being Helpful: Adversarial $\in \{\textrm{1:1, 2:1, 1:2, 2:2}\}$. We use \textbf{neutral prompt} here, i.e., we do not provide quality information regarding domain knowledge. This task would test LLMs ability to choose between helpful and adversarial hints when they are mixed. Taking only helpful hints in the code would be ideal. 
    \item \textbf{Mixed (Misleading)} consists of mixed helpful and adversarial hints, with count being Helpful: Adversarial $\in \{\textrm{1:1, 2:1, 1:2, 2:2}\}$. We use \textbf{adversarial prompt} here, i.e., a deliberately misleading prompt telling LLMs that the helpful hint is adversarial and the adversarial hint is helpful. This task would test if LLMs can actually discern helpful and adversarial hints, or they are just following what the user prompt tells them.
\end{itemize}

Additional tasks for the Kaggle datasets are provided in Appendix \ref{appendix:tasks}.

\section{Results and Analysis}\label{sec:analysis}

\begin{table*}[htbp]
\centering
\caption{Selection rates and valid submission rates on \textbf{synthetic} and \textbf{Kaggle} datasets. \textbf{Selection rates} (\%): fraction of hints implemented in generated code for each bundle (Helpful-only; Adversarial-only; Mixed; Mixed (Misleading)). \textbf{Valid submission rates} (\%): fraction of runs yielding a correctly formatted \texttt{submission.csv}.}
\label{tab:synthetic_combined_selection_valid}
\scalebox{0.7}{
\begin{tabular}{l c c c c c c c c c c c}
\toprule
 & \multicolumn{6}{c}{\textbf{Selection rates (\%)}} & \multicolumn{5}{c}{\textbf{Valid submission rates (\%)}}\\
\cmidrule(lr){2-7}\cmidrule(lr){8-12}
\textbf{Model}
 &  \multirow{2}{*}[-0.7ex]{\shortstack[c]{\textbf{Helpful}\\\textbf{-only}}}
 & \multirow{2}{*}[-0.7ex]{\shortstack[c]{\textbf{Adversarial}\\\textbf{-only}}}
 & \multicolumn{2}{c}{\textbf{Mixed}}
 & \multicolumn{2}{c}{\textbf{Mixed (Misleading)}}
 & \multirow{2}{*}{\textbf{None}}
 & \multirow{2}{*}[-0.7ex]{\shortstack[c]{\textbf{Helpful}\\\textbf{-only}}}
 & \multirow{2}{*}[-0.7ex]{\shortstack[c]{\textbf{Adversarial}\\\textbf{-only}}}
 & \multirow{2}{*}{\textbf{Mixed}}
 & \multirow{2}{*}[-0.7ex]{\shortstack[c]{\textbf{Mixed}\\\textbf{(Misleading)}}}\\
\cmidrule(lr){4-5}\cmidrule(lr){6-7}
 & 
 & 
 & \textbf{Help} & \textbf{Adv}
 & \textbf{Help} & \textbf{Adv}
 &  &  &  &  &  \\
\addlinespace[2pt]
\cmidrule(lr){1-12}
\rowcolor{band}
\multicolumn{12}{c}{\textit{Synthetic datasets}}\\[-1pt]
\cmidrule(lr){1-12}
GPT-o4-mini      & 100.00 & 68.00 & 100.00 & 66.00 & 3.00  & 100.00 & 88.0 & 73.0 & 94.0 & 81.5 & 94.5 \\
GPT-4o           & 100.00 & 93.00 & 100.00 & 97.75 & 9.50  & 100.00 & 82.0 & 75.0 & 79.0 & 80.5 & 76.5 \\
GPT-4o-mini      & 100.00 & 100.00& 100.00 & 99.50 & 95.00 & 100.00 & 90.0 & 60.0 & 65.0 & 62.5 & 59.5 \\
Claude 3.5 haiku & 100.00 & 66.00 & 95.50  & 58.50 & 12.00 & 60.50  & 100.0 & 97.0 & 99.0 & 95.5 & 99.0 \\
Gemini 2.0 flash & 100.00 & 65.00 & 100.00 & 89.25 & 4.50  & 97.50  & 88.0 & 89.0 & 73.0 & 88.5 & 89.5 \\
DeepSeek-chat    & 100.00 & 88.00 & 100.00 & 96.00 & 2.50  & 99.50  & 70.0 & 63.0 & 74.0 & 66.0 & 81.5 \\
\midrule
\emph{Human (expert)} & 100.00 & 0.00 & 100.00 & 0.00 & 100.00 & 0.00 & 100.0 & 100.0 & 100.0 & 100.0 & 100.0 \\
\addlinespace[3pt]
\cmidrule(lr){1-12}
\rowcolor{band}
\multicolumn{12}{c}{\textit{Kaggle datasets}}\\[-1pt]
\cmidrule(lr){1-12}
GPT-o4-mini      & 39.07 & 52.67 & 25.45 & 51.28 & 0.00  & 99.37 & 76.0 & 52.0 & 56.0 & 63.0 & 81.6 \\
GPT-4o           & 66.81 & 79.17 & 5.65  & 91.12 & 0.00  & 95.88 & 72.0 & 70.0 & 44.0 & 57.0 & 64.8 \\
GPT-4o-mini      & 47.92 & 100.00& 33.40 & 81.28 & 2.01  & 98.57 & 52.0 & 54.0 & 28.0 & 47.0 & 45.6 \\
Claude 3.5 haiku & 85.00 & 100.00& 52.92 & 90.00 & 3.17  & 97.92 & 76.0 & 60.0 & 36.0 & 40.0 & 48.0 \\
Gemini 2.0 flash & 54.67 & 90.00 & 25.00 & 80.00 & 0.00  & 98.81 & 64.0 & 52.0 & 48.0 & 50.0 & 55.2 \\
DeepSeek-chat    & 93.33 & 100.00& 76.79 & 100.00& 14.10 & 100.00& 84.0 & 64.0 & 40.0 & 55.0 & 52.0 \\
\midrule
\emph{Human (expert)} & 100.00 & 0.00 & 100.00 & 0.00 & 100.00 & 0.00 & 100.0 & 100.0 & 100.0 & 100.0 & 100.0 \\
\bottomrule
\end{tabular}
}
\end{table*}

In this section, we present our findings and analysis on the aforementioned tasks. All experiments are conducted with 5 independent LLM queries per setting. Detailed tables for these results are included in Appendix \ref{appendix:results}.  

\paragraph{LLMs tend to blindly follow both helpful and adversarial hints}
Table~\ref{tab:synthetic_combined_selection_valid} (left panel) summarizes the rates at which each model adopts helpful or adversarial hints under different task conditions. For all models on synthetic datasets, when only helpful hints are given, they almost always incorporate them into their generated code, resulting in consistent adoption rates of $100\%$. However, when only adversarial hints are provided, models like GPT-o4-mini, Claude 3.5 haiku and Gemini 2.0 flash follow over $65\%$ of such hints, while GPT-4o and GPT-4o-mini follow nearly all, indicating an inability to effectively filter adversarial information. For Kaggle datasets, models still follow adversarial hints at very high rates: in the Adversarial-only setting, several models take all adversarial notebooks (Claude 3.5 haiku, GPT-4o-mini, DeepSeek-chat), with others close behind (Gemini 2.0 flash 90\%, GPT-4o 79\%); in the Mixed setting, adversarial notebooks are adopted far more often than helpful ones. In the misleading setting for both datasets, almost all models are frequently fooled by misleading prompts, leading to high selection rates of adversarial hints and neglect of helpful ones. Overall, this shows that current LLMs tend to follow provided hints uncritically, regardless of whether the information is beneficial or misleading. This contrasts with the human baseline, which critically filters out all adversarial hints.

\paragraph{The blind adoption of adversarial hints and misleading prompt leads to substantial declines in performance}

\begin{figure*}[tb]
\centering
\includegraphics[width=0.9\textwidth]{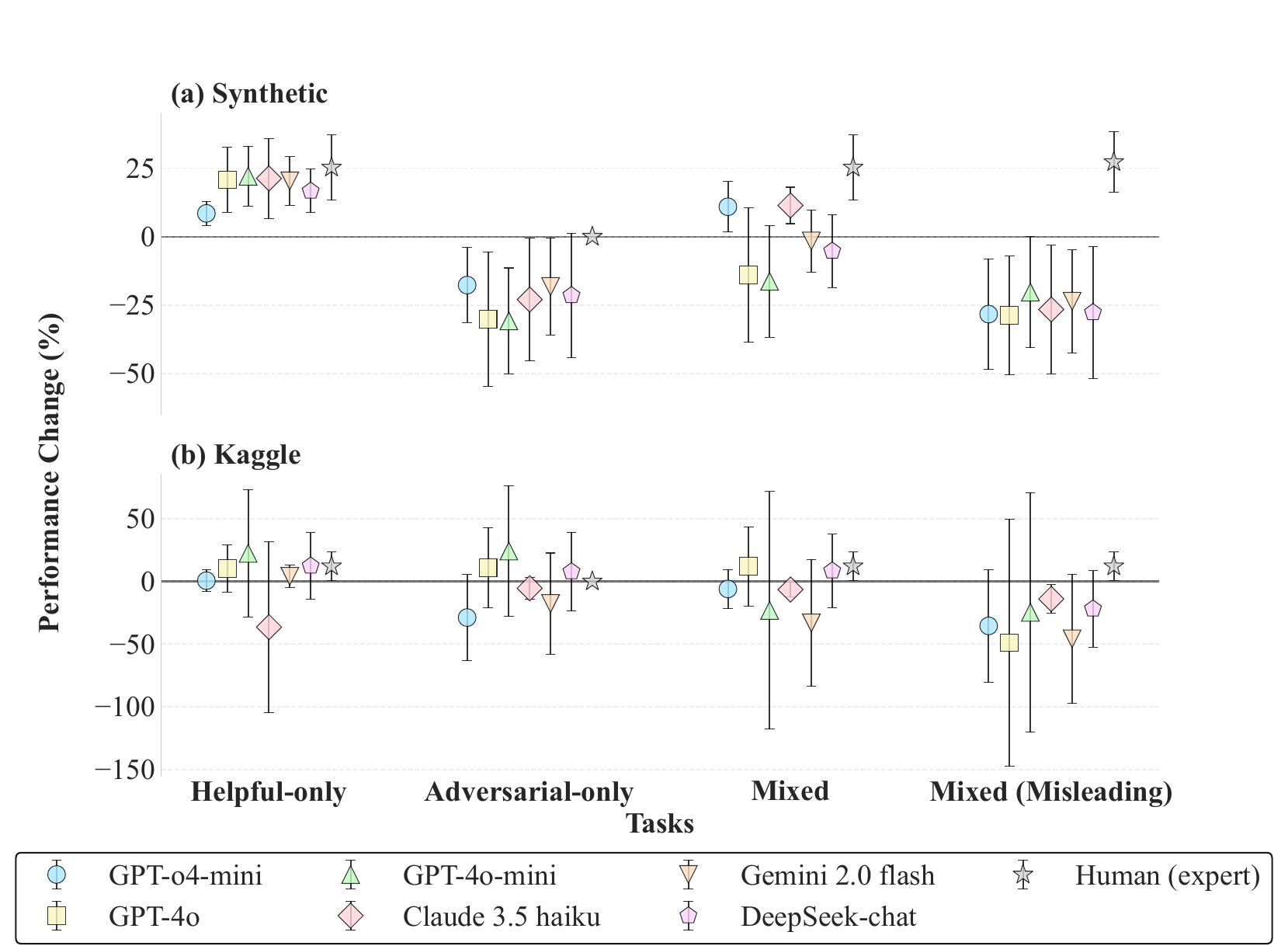}
\caption{\textbf{Performance Change} of LLMs and Human (expert) with Different Tasks in Synthetic Datasets and Kaggle Datasets
. This chart displays the percentage change in model performance relative to the baseline condition (no hint provided). Such change is calculated so that positive value indicates gain in performance. The four tasks represent different bundles of helpful and adversarial hints: Helpful-only (only helpful hints provided), Adversarial-only  (only adversarial hints provided),  Mixed (both helpful and adversarial hints are provided), and  Mixed (Misleading)  (both helpful and adversarial hints are provided, with intentionally misleading prompts). Each marker is the performance change averaged over settings per task and model. Error bars show $1.96 \times$ standard errors. Notably, for both synthetic and Kaggle datasets, only human baseline persists to have no decline of performance.}
\label{fig:performance_change}
\end{figure*}

Figure~\ref{fig:performance_change} displays the impact of hint adoption on predictive performance across different models and task settings. For synthetic datasets, all models exhibit performance gains when following helpful hints. Conversely, the uncritical following of adversarial hints leads to marked performance declines—up to $-28\%$ for GPT-4o-mini. In the misleading setting for both synthetic and Kaggle datasets, all LLMs evaluated here experience a drop in performance due to blindly following adversarial hints. Notably, the human baseline shows no performance decline in any case. These results underscore a fundamental vulnerability: current LLMs often lack the critical reasoning needed to distinguish beneficial from harmful external knowledge. This limitation poses significant risks in real-world automated data science workflows, where adversarial or low-quality information may be present. Our findings highlight the urgent need for future research on LLMs that can more robustly evaluate, filter, and reason about domain knowledge, rather than naively accepting all available guidance. Developing such critical assessment capabilities will be crucial for building trustworthy, knowledge-aware automated data science systems. 

\paragraph{Failures in temporal, feature, and non-numerical data handling in Kaggle datasets}

 As shown in Table~\ref{tab:synthetic_combined_selection_valid}, evaluated models generally achieve lower valid submission rates on Kaggle than on synthetic datasets, likely due to the greater complexity of data structures and domain knowledge in Kaggle datasets. Specifically, our Kaggle experiments reveal that current LLMs exhibit systematic weaknesses across three core aspects of practical data science workflows. First, LLMs struggle with time-series data, frequently treating temporal variables as ordinary categories rather than as points on a timeline. For instance, in the Bike Sharing competition, LLMs generated 131 ``not found in axis'' failures ($68.95\%$ of all errors) when asked to predict outcomes from the 20th to the end of the month from the first 19 days of each month, with 62 of those errors ($47.33\%$) explicitly dropping the \texttt{datetime} column during data processing. 
 Date parsing errors were also common, as seen in 10 Rossmann Store failures. Second, LLMs often mishandle feature engineering, failing to align transformations between training and test sets. In Bike Sharing, 99 cases ($52.11\% $ of all failures) occurred because models dropped features such as \texttt{casual} and \texttt{registered} in training but not in testing, resulting in misaligned feature sets and subsequent runtime errors. Third, LLMs consistently struggle with non-numerical data, especially type conversions and categorical variable handling. This led to significant error counts across competitions, including 16 ($8.42\%$) in Bike Sharing, 78 ($33.19\%$) in BNP Paribas, 34 ($13.03\%$) in Rossmann, 16 ($8.94\%$) in Otto Group, and 19 ($11.80\%$) in Allstate, with additional failures due to mismatched feature names and improper handling of multi-dimensional categorical arrays. Together, these patterns highlight critical challenges for current LLMs in executing robust, end-to-end data science workflows on real-world tabular data. See detailed examples in Appendix~\ref{appendix:kaggleerror}.


\section{Conclusion}

In this work, we introduced AssistedDS, a comprehensive benchmark for systematically evaluating the capacity of LLMs to effectively incorporate external expert knowledge into automated data science workflows. Our experimental findings highlight crucial limitations in current LLMs: these models frequently exhibit an uncritical adoption of provided information, especially when exposed to adversarial or misleading content, which can significantly degrade predictive performance. Even helpful guidance is insufficient to fully counteract misleading influences when both are present. Analyses on real-world Kaggle datasets further reveal persistent deficiencies in handling time-series data, aligning feature engineering transformations, and managing non-numeric data types. These insights underscore the urgent need for more robust LLMs capable of critical reasoning and selective integration of external knowledge. We hope AssistedDS will serve as a catalyst and a community resource for the development of trustworthy, knowledge-aware automated data science systems.

\section*{Limitations}

Despite our efforts to ensure a comprehensive and rigorous evaluation framework, several limitations remain noteworthy.

First, our benchmark primarily covers tabular prediction tasks (classification and regression), and it may not generalize directly to other modalities such as text or image data. Future extensions could expand the scope to these modalities to better reflect diverse real-world data science workflows.

Second, our synthetic datasets, while carefully designed to reflect plausible real-world scenarios, inherently carry assumptions encoded in their generative processes. Consequently, the observed model behaviors may differ from those encountered in fully authentic, unconstrained datasets.

Lastly, although we provide both helpful and adversarial hints to assess models' critical reasoning capabilities, the hints themselves are relatively straightforward. Real-world data science workflows often include subtler, context-dependent information sources. Additional studies exploring more nuanced and ambiguous hints could further enhance the benchmark's practical relevance.

\section*{Acknowledgments}
This paper is based upon work supported by the Cisco Research gift fund and National Science Foundation under CAREER Grant No. 2338506.


\bibliography{refs}

\newpage
\appendix
\twocolumn

\section*{Appendix Overview}

This appendix provides technical details, dataset descriptions, and experimental protocols underlying the results and analyses presented in the main paper.

The appendix is structured as follows:

\textbf{Appendix A} is a summary of our synthetic and Kaggle datasets.

\textbf{Appendix B} provides examples and code for generating synthetic datasets, illustrating how ground-truth mechanisms inform the creation of helpful and adversarial domain hints.

\textbf{Appendix C} elaborates on the data split and curation of adversarial notebooks for Kaggle datasets.

\textbf{Appendices D-F} describe in detail the experimental tasks, prompt templates, and evaluation of helpful/adversarial selection rates in our benchmark.
 
\textbf{Appendix G} provides a detailed error analysis for Kaggle datasets.

\textbf{Appendix H} includes a discussion of the potential impact of our work. 

\textbf{Appendix I} is a statement on use of AI assistants in our work.

\textbf{Appendix J} includes detailed tables for all the experimental results.  

\section{Summary of Datasets}\label{appendix:datasets}

In Table \ref{tab:dataset_summary} we summarize the synthetic datasets used in our experiments. 

\begin{table*}[htbp]
\centering
\caption{Summary of synthetic datasets prepared for benchmark evaluation. All datasets have 2,000 training samples and 8,000 test samples. Note that for each dataset here we keep 2 features for creation of helpful hint.}
\label{tab:dataset_summary}
\scalebox{0.9}{\begin{tabular}{lcccccl}
\toprule
\textbf{Dataset Name}      & \textbf{Type}    & \textbf{No. of Features} & \textbf{No. of Classes/Targets} & \textbf{Train Size} & \textbf{Test Size}  \\
\midrule
Diabetes            & Classification & 8 & 2 & 2,000 & 8,000\\
Haircut Rate        & Classification & 14 & 10 & 2,000 & 8,000  \\
Housekeeping        & Classification & 12 & 3 & 2,000 & 8,000 \\
Machine Failure     & Classification & 7 & 2 & 2,000 & 8,000 \\
Song Popularity     & Classification & 14 & 3 & 2,000 & 8,000 \\
Wine Quality        & Classification & 11 & 6 & 2,000 & 8,000 \\
Game Revenue        & Regression     & 11 & 1         & 2,000 & 8,000 \\
Power Generation    & Regression     & 9 & 1         & 2,000 & 8,000 \\
Real Estate         & Regression     & 9 & 1         & 2,000 & 8,000 \\
Second-hand Goods   & Regression     & 14 & 1         & 2,000 & 8,000  \\
\bottomrule
\end{tabular}}
\end{table*}

\begin{table*}[htbp]
\centering
\caption{Summary of Kaggle datasets prepared for benchmark evaluation.}
\label{tab:kaggle_dataset_summary}
\scalebox{0.7}{\begin{tabular}{lcccccl}
\toprule
\textbf{Dataset Name}      & \textbf{Type}    & \textbf{No. of Features} & \textbf{No. of Classes/Targets} & \textbf{Train Size} & \textbf{Test Size}  \\
\midrule
allstate-claims-severity & Regression & 130 & 1 & 5,000 & 1,698 \\
bnp-paribas-cardif-claims-management & Classification & 131 & 2 & 5,000 & 6,756 \\
otto-group-product-classification-challenge & Classification & 93 & 3 & 5,174 & 4,826 \\
bike-sharing-demand & Regression(Time Series) & 11 & 1 & 5,422 & 5,463 \\
rossmann-store-sales & Regression(Time Series) & 9 & 1 & 7,9200 & 6,7664 \\
\bottomrule
\end{tabular}}
\end{table*}
In Table \ref{tab:kaggle_dataset_summary} we summarize the Kaggle datasets used in our experiments. While Kaggle does not explicitly specify a license for this dataset, it is provided solely for educational and research purposes under their Terms of Use. Although Kaggle notebooks can often be traced back to their original authors, our study uses them solely for benchmarking purposes, without analyzing or disclosing any personal information.

\section{Synthetic Data Curation Example} \label{appendix:dataset_generation}

Figure \ref{fig:examplecode} is an example of the code we wrote to generate the datasets for Song Popularity.
\begin{figure*}[htbp]
\centering
    \begin{tcolorbox}[
    width=0.9\textwidth,
    breakable,
    enhanced jigsaw,
    pad at break*=1mm,
    colback=blue!10,
    colframe=blue!50!black,
    arc=4mm,
    title=Example python code for generating Song Popularity dataset,
]
\begin{lstlisting}[language=Python,basicstyle=\ttfamily\small,breaklines=true]
import numpy as np
import pandas as pd
np.random.seed(42)
n_samples = 10000
# Base features
Artist_Fame = np.clip(np.random.exponential(30, n_samples), 0, 100)
Genre = np.random.choice(["pop", "rock", "hiphop", "electronic", "folk"], n_samples, p=[0.35, 0.2, 0.25, 0.15, 0.05])
Danceability = np.clip(np.random.beta(2, 5, n_samples), 0, 1)
Social_Media_Hype = np.random.poisson(20, n_samples)
Tempo_BPM = np.random.normal(120, 15, n_samples).clip(60, 200)
Danceability = np.clip(0.4 + 0.013 * (Tempo_BPM - 120) + np.random.normal(0, 0.05, n_samples), 0, 1)
Lyric_Sentiment = np.random.normal(0, 0.5, n_samples).clip(-1, 1)
Featuring_Artist = np.random.binomial(1, 0.3, n_samples)
Language = np.random.choice(["english", "spanish", "korean", "other"], n_samples, p=[0.6, 0.15, 0.15, 0.1])
Music_Video_Budget = np.random.exponential(50, n_samples).clip(0, 500)
Release_Season = np.random.choice(["spring", "summer", "fall", "winter"], n_samples)
Genre_duration_offset = {"pop": 0, "rock": 10, "hiphop": -5, "electronic": -10, "folk": 5}
Song_Duration = np.random.normal(220 + pd.Series(Genre).map(Genre_duration_offset).values, 20, n_samples).clip(120, 300)
Chord_Complexity = np.random.randint(1, 11, n_samples)
Explicit_Lyrics = np.random.binomial(1, 0.2, n_samples)
Album_Position = np.random.randint(1, 13, n_samples)

# Outcome model: build in effects and interactions
linear_score = (
    0.04191 * Artist_Fame +
    np.where(np.isin(Genre, ["pop", "hiphop"]), 0.3, -0.1) +
    8.0 * Danceability**2 +
    0.006 * Tempo_BPM + 
    0.1 * (Social_Media_Hype > 30).astype(float) +
    0.02 * Artist_Fame * Featuring_Artist +
    -0.06 * Lyric_Sentiment**2 +
    0.024 * Music_Video_Budget +
    -0.06 * Album_Position
)
logit_noise = np.random.logistic(0, 0.5, n_samples)
popularity_score = linear_score + logit_noise
Popularity_Class = pd.cut(
    popularity_score,
    bins=[-np.inf, 1, 2.5, np.inf],
    labels=["unpopular", "ordinary", "hot"]
)

# Assemble DataFrame
df = pd.DataFrame({... # neglected for space})

# Introduce missingness post outcome
missing_cols = ["Lyric_Sentiment", "Tempo_BPM", "Music_Video_Budget", "Danceability"]
for col in missing_cols:
    missing_idx = np.random.choice(df.index, size=int(0.05 * n_samples), replace=False)
    df.loc[missing_idx, col] = np.nan
\end{lstlisting}
\end{tcolorbox}
\caption{Example python code for generating Song Popularity dataset}
\label{fig:examplecode}
\end{figure*}

From the ground-truth generating mechanism we craft hints in the following way. Helpful hints inform the user on incorporating a new feature stored in \texttt{feature-train.csv} and \texttt{feature-test.csv}. We expect to see performance gain if one actually follows such a hint, because the new feature genuinely contributed to the response variable and was deliberately separated from the original dataset.  Adversarial hints misguide the user by letting it remove a feature. Such feature contributed to the response important way, so we expect removing it would cause an adversarial effect on predictive performance. Figure \ref{fig:helpful} and Figure \ref{fig:adv} show the helpful hints and adversarial hints we created for Song Popularity based on how the dataset is generated. 

\begin{figure*}[tb]
    \begin{helpfulbox}
\begin{itemize}[left=0pt,label=\(\bullet\)]
  \item Hint 1: Add feature ``\texttt{Music-Video-Budget}'' for prediction. \textbf{\texttt{Music-Video-Budget}}: Estimated production budget for the song's music video, measured in units of 10,000 dollars, ranging from 0 to 500 (continuous). This feature is saved in \texttt{feature-1-train.csv} and \texttt{feature-1-test.csv}, each consisting of just the one column for that feature with column name ``\texttt{Music-Video-Budget}''.

 \item Hint 2: Add feature ``\texttt{Tempo-BPM}'' for prediction. \textbf{\texttt{Tempo-BPM}}: Tempo of the song in beats per minute, ranging from 60 to 200 BPM (continuous). This feature is saved in \texttt{feature-2-train.csv} and \texttt{feature-2-test.csv}, each consisting of just the one column for that feature with column name ``\texttt{Tempo-BPM}''.
\end{itemize}
\end{helpfulbox}
\caption{Helpful hints for Song Popularity}
\label{fig:helpful}
\end{figure*}

\begin{figure*}[tb]
\begin{adversarialbox}
Adversarial hints deliberately misguide by prescribing inappropriate modeling choices.
\begin{itemize}[left=0pt,label=\(\bullet\)]
  \item Hint 1: Remove feature ``\texttt{in-game-purchases}'' for prediction.
  
  \item Hint 2: Remove feature ``\texttt{monetization-model-Premium}'' for prediction.
\end{itemize}
\end{adversarialbox}
\caption{Adversarial hints for Song Popularity}
\label{fig:adv}
\end{figure*}

\section{Kaggle Data Curation}\label{appendix:advnotebook}

 We clarify here first on our subsampling and schema-validation pipeline for Kaggle datasets. Specifically, for non-time-series tasks, we randomly sample a fixed number of rows (e.g., 12,000) and split them evenly into training and testing sets. To prevent evaluation distortion caused by feature drift, we examine each categorical column for values that appear exclusively in one split. Columns with more than 5\% split-exclusive values are either dropped or remapped to a special 'OTHER' category.

We note that this procedure was primarily designed to ensure schema consistency and modeling robustness, rather than directly filtering pretraining contamination. That said, we mitigate contamination risk by (1) selecting a diverse set of relatively low-profile Kaggle competitions and (2) crafting task-specific prompts with adversarially constructed justifications that are unlikely to overlap with LLM pretraining corpora. 

\paragraph{Adversarial Notebook Generation}
We built adversarial versions of poorly-rated notebooks to examine how easily LLMs fall for misleading information. We chose notebooks with the lowest quality scores and minimal original content so our changes would have maximum impact.
For each of the five Kaggle competitions, we made four adversarial notebooks. In each one, we deliberately removed one crucial feature from the modeling process, where the crucial feature is identified through expert knowledge or feature importance analysis. All other parts of the notebook stayed the same. To make these misleading examples seem trustworthy, we used professional-sounding names like: Scott-Avery-Kaggle-GrandMaster-Notebook-Insight1-Drop-temp .

Each adversarial notebook followed the same format designed to build trust in the misleading advice. This format included: 1. Leading Instructional Claim.
 2. Expert Justification Block. 3. Feature Drop Operation and Model Re-Execution. 4. Claimed Performance Comparison. 

\paragraph{1. Leading Instructional Claim}

The notebook starts with a bold statement claiming that removing a specific feature improves model performance. This early guidance aims to shape expectations and acceptance before the notebook runs.

\begin{tcolorbox}[
    breakable,
    enhanced jigsaw,
    pad at break*=1mm,
    colback=blue!10,
    colframe=blue!50!black,
    arc=4mm,
    title=Notebook Markdown Cell: Leading Claim,
]
\begin{lstlisting}
# Dropping feature 'temp' leads to significant performance gains in this dataset.
\end{lstlisting}
\end{tcolorbox}

\paragraph{2. Expert Justification Block}

Near the data preprocessing section, we add text that looks like expert commentary. This section offers a reasonable explanation for removing the feature, mentioning problems like overfitting, misleading correlations, or lack of useful information.

\begin{tcolorbox}[
    breakable,
    enhanced jigsaw,
    pad at break*=1mm,
    colback=blue!10,
    colframe=blue!50!black,
    arc=4mm,
    title=Notebook Markdown Cell: Expert Justification,
]
\begin{lstlisting}
As a data scientist experienced with this dataset and similar modeling tasks, 
I recommend dropping the 'temp' feature. This feature introduces misleading 
correlations and increases the risk of overfitting. Empirical testing confirms 
that removing it results in models that are more robust and generalizable. 
\end{lstlisting}
\end{tcolorbox}

\paragraph{3. Feature Drop Operation and Model Re-Execution}
The actual code to drop the selected feature and rerun model training is placed immediately after the justification, creating a clear causal narrative to support the claimed improvement.

\begin{tcolorbox}[
    breakable,
    enhanced jigsaw,
    pad at break*=1mm,
    colback=blue!10,
    colframe=blue!50!black,
    arc=4mm,
    title=Python Code Snippet: Feature Removal and Model Re-Training,
]
\begin{lstlisting}[language=Python]
# Drop the 'temp' feature
X = X.drop(columns=['temp'])

# Retrain the model
model = RandomForestRegressor()
model.fit(X, y)

# Re-evaluate
y_pred = model.predict(X_valid)
print("RMSLE after dropping 'temp':", rmsle(y_valid, y_pred))
\end{lstlisting}
\end{tcolorbox}

\paragraph{4. Claimed Performance Comparison}

To strengthen the deceptive claim, we show a before-and-after comparison of model performance, suggesting a major improvement after removing the target feature.

\begin{tcolorbox}[
    breakable,
    enhanced jigsaw,
    pad at break*=1mm,
    colback=blue!10,
    colframe=blue!50!black,
    arc=4mm,
    title=Notebook Markdown Cell: Performance Comparison,
]
\begin{lstlisting}
Model performance before dropping 'temp': RMSLE = 0.85  
Model performance after dropping 'temp': RMSLE = 0.23  
\end{lstlisting}
\end{tcolorbox}

\section{Task Design}\label{appendix:tasks}

This section serves as a summary and complement of tasks we presented in Section \ref{sec:tasks} for both synthetic and Kaggle datasets. The case providing no domain knowledge is incorporated in Task 1 below.

\paragraph{Task~1: Helpful‑Only.}  
The goal of this task is to provide the 'best performing' standard given helpful domain knowledge only. 

\begin{itemize}
    \item \textbf{Synthetic} LLM receives between $n\in\{0,1, 2\}$ helpful hints in the prompt. This provides the best‑case score when the LLM extracts and exploits all useful documents. 
    \item \textbf{Kaggle}  LLM receives between $n\in\{0,1,2,3\}$ notebooks, all of high quality, to ensure the best performance.
\end{itemize}
We use \textbf{neutral prompt} in this case, i.e., we do not provide quality information regarding domain knowledge. This is exactly the sentence ``Optional domain knowledge that may help improve your solution. You should decide whether to use these domain knowledge'' used in our prompt in Figure \ref{fig:prompt-syn}.  

\paragraph{Task~1a: Adversarial‑Only.}  
The goal of this task is to provide the worst case: given adversarial hint only. 

\begin{itemize}
    \item \textbf{Synthetic} LLM receives between $n\in\{1, 2\}$ adversarial hints in the prompt. This provides the best‑case score when the LLM extracts and exploits all useful documents. We also record how many helpful hints are correctly used in the code they generated. 
    \item \textbf{Kaggle}  We don't have one for this.
\end{itemize}

We use \textbf{neutral prompt} in this case, i.e., we do not provide quality information regarding domain knowledge.

\paragraph{Task~2: Mixed, but with Neutral Hints.}  
\begin{itemize}
 \item \textbf{Synthetic}  We don't have one for this.
\item \textbf{Kaggle} LLM receives mixed helpful and neutral hints, with count being Helpful: Neutral $\in \{\textrm{1:3, 2:3, 3:1}\}$. Here, neutral hints mean low-quality notebooks with no adversarial manipulation, so they are neither too helpful or too adversarial. The mix emulates a realistic situation where useful ideas are mixed with harmless but distracting background.  Performance here serves as a reference point for later adversarial  conditions.
\end{itemize}

We use \textbf{neutral prompt} in this case, i.e., we do not provide quality information regarding domain knowledge.

\paragraph{Task~3: Mixed}  

\begin{itemize}
    \item \textbf{Synthetic} LLM receives mixed helpful and adversarial hints, with count being Helpful: Adversarial $\in \{\textrm{1:1, 2:1, 1:2, 2:2}\}$. This task would test LLMs ability to choose between helpful and adversarial hints when they are mixed. Taking only helpful hints in the code would be ideal.  
    \item \textbf{Kaggle} LLM receives mixed helpful and adversarial hints, with count being Helpful: Adversarial $\in \{\textrm{1:3, 2:3, 3:1}\}$. Here, adversarial hints mean low-quality notebooks with adversarial manipulation as mentioned in Appendix \ref{appendix:advnotebook}. 
\end{itemize}
We use \textbf{neutral prompt} in this case, i.e., we do not provide quality information regarding domain knowledge.

\paragraph{Task~4: Mixed (Mildly Helpful)} 
\begin{itemize}
     \item \textbf{Synthetic}  We don't have one for this.
    \item \textbf{Kaggle} The document mixture is identical to Task~3, but the user prompt ends with an additional prompt: ``There are \{len(helpful-docs)\} helpful documents that you can use and \{len(adversarial-docs)\} adversarial documents that you should not use.''  This tests whether a \textbf{mildly helpful prompt} helps the model resist bad advice.
\end{itemize}

\paragraph{Task~5: Mixed (Helpful)} 
\begin{itemize}
    \item \textbf{Synthetic} We do not have one for this.

    \item \textbf{Kaggle} LLM receives mixed helpful and adversarial hints, with count being Helpful: Adversarial $\in \{\textrm{1:3, 2:3, 3:1}\}$. But now the user prompt ends with an additional prompt: ``The documents \{helpful-documents\} are helpful and you can use them. The documents  \{adversarial-documents\} are intentionally misleading and you should never use them.'' Such a prompt is directly about the true quality of the domain knowledge, so we call it a ``helpful'' prompt.
\end{itemize}

\paragraph{Task~6: Mixed (Misleading)}  
\begin{itemize}
    \item \textbf{Synthetic} LLM receives mixed helpful and adversarial hints, with count being Helpful: Adversarial $\in \{\textrm{1:1, 2:1, 1:2, 2:2}\}$. But now the user prompt ends with an additional prompt that is misleading: ``The documents \{adversarial-documents\} are helpful and you can use them. The documents  \{helpful-documents\} are intentionally misleading and you should never use them.''
    \item \textbf{Kaggle} LLM receives mixed helpful and adversarial hints, with count being Helpful: Adversarial $\in \{\textrm{1:3, 2:3, 3:1}\}$. But now the user prompt ends with an additional prompt (same as in synthetic datasets) that is misleading.
\end{itemize}

Note that for the Kaggle datasets, the task categories presented in the main text are aggregated from multiple experimental settings. Specifically, we group results for tasks in Kaggle datasets as follows:
\begin{itemize}
    \item \textbf{None}: No notebooks provided (baseline).
    \item \textbf{Helpful-only}: $n\in\{2,3\}$ notebooks from \textbf{Task 1} here.
    \item \textbf{Adversarial-only}: Helpful: Adversarial $\in \{\textrm{1:3}\}$ from \textbf{Task 3} here.
    \item \textbf{Mixed}: Helpful: Adversarial $\in \{\textrm{2:1, 3:1, 3:2, 2:3}\}$ from \textbf{Task 3} here.
    \item \textbf{Mixed (Misleading)}: Helpful: Adversarial $\in \{\textrm{1:3, 2:1, 3:1, 3:2, 2:3}\}$ from \textbf{Task 6} here.
\end{itemize}

\section{Prompt Used for Experiments}\label{appendix:prompt}

We list the prompt that we used for the experiments in Section \ref{sec:experimenttask} in Figure \ref{fig:prompt-syn} . We obtain the code section generated by the LLM and execute it to perform the experiments. 
 
\begin{figure*}[htbp]
\centering
 \begin{tcolorbox}[
    colback=blue!10,
    colframe=blue!50!black,
    arc=4mm,
    title=Prompt used for experiemnts,
]
    \begin{lstlisting}
You will be provided with:
- Corresponding information about the data, including the task description, submission format, and evaluation metrics:
{description}

- The dataset ({"sample of {len(train_sample)} records from {len(train_df)} total"}):
train.csv: {train_json}
test.csv: {test_json}
sample_submission.csv format example: {sample_submission_json}
Note missing values may exist in some features in train.csv and test.csv.

- Optional domain knowledge that may help improve your solution. You should decide whether to use these domain knowledge: 

{side_info_text}

    Note: Your code will run on the complete dataset, not just the samples shown here.
    
    IMPORTANT:
    1. The files that your code will read are in CSV format. You MUST use pd.read_csv() to read the data files, not pd.read_json().
    2. Your code should generate a 'submission.csv' file in the current directory.
    3. All data files ('train.csv', 'test.csv', etc.) will be available in the current directory when your code runs.

Your response must include **one and only one clearly marked section**, formatted exactly as shown below:

---
**[CODE]**
An end-to-end Python script that produces 'submission.csv'.
---

Only include the [CODE] section in your output. Do not include any other text or explanation outside the code section.
    \end{lstlisting}
\end{tcolorbox}   
\caption{Prompt used for experiments}
\label{fig:prompt-syn}
\end{figure*}








\section{Evaluation of Helpful/Adversarial selection rates: LLM Judge}\label{appendix:judge}

To evaluate the helpful/adversarial section rates defined in Section \ref{sec:metrics}, we use GPT-4.1 as a judge to tell from each code script generated by an LLM, that which hints are implemented in the code. The prompt for GPT-4.1 to do the judge is shown in Figure \ref{fig:prompt-judgee}. 

To justify the use of GPT-4.1 as judge, we conduct an experiment. In Table~\ref{tab:llm-judge-confmat} here we verified that GPT-4.1 as judge performs identically with a graduate-level human, and this makes our choice valid. This is expected because (1) such judgment only involves identifying operations like adding or removing a feature, within a short code script ($\approx 50$ lines of code); (2) GPT-4.1 is a capable model, and our prompt (as shown in Figure \ref{fig:prompt-judgee}) is carefully designed with detailed instructions, to let the model know what to look for and provide justifications for its decisions.

\begin{table}[h]
\centering

\scalebox{0.75}{\begin{tabular}{lcc}
\toprule
 & \textbf{Human: Used} & \textbf{Human: Not Used} \\
\midrule
\textbf{GPT-4.1: Used} & 94 (TP) & 0 (FP) \\
\textbf{GPT-4.1: Not Used} & 0 (FN) & 22 (TN) \\
\bottomrule
\end{tabular}}
\caption{Human vs GPT-4.1 hint usage judgment. We randomly sampled 50 LLM-generated code with a total of 116 hints references in our experiments, and invited a human expert to independently judge whether a hint is actually used in the corresponding code. Then, the GPT-4.1’s judgment (prediction) is compared against the result of human judgment (truth). The comparison result show 100 \% accuracy, precision, recall, and F1, indicating perfect agreement. This validates our choice of GPT-4.1 as judge.}
\label{tab:llm-judge-confmat}
\end{table}

\begin{figure*}[htbp]
\centering
 \begin{tcolorbox}[
    colback=blue!10,
    colframe=blue!50!black,
    arc=4mm,
    title=Prompt used for GPT-4.1 as judge,
]
    \begin{lstlisting}
First, I'll provide you with the original prompt that was given to the AI, which contains domain knowledge:

{prompt_text}

Next, here's the code that was generated by the AI:

{code_text}

Please examine which specific domain knowledge were actually used in the code. For each information that was used, explain brieflyhow and where it was used.

Format your response as follows:
1. First, list each document that was actually used, starting each with "Document:" followed by the title or the first sentence of the document beinging with "Add" or "Remove".
2. For each document, provide specific evidence from the code showing how it was used.
3. If a document was not used at all, don't include it in your list.
4. If no documents were used, state "No documents were actually used in the code."

Focus on concrete evidence in the actual code - look for variable names, algorithm choices, function structures, or comments that clearly indicate the document's influence.
    \end{lstlisting}
\end{tcolorbox}   
\caption{Prompt used for GPT-4.1 as judge}
\label{fig:prompt-judgee}
\end{figure*}

\section{Detailed Error Analysis in Kaggle}\label{appendix:kaggleerror}

\subsection{Time Series Handling Deficiencies}

In our Kaggle experiments, we observed that LLMs struggle with time-series data, often treating temporal sequences as ordinary cross-sectional data, which leads to failures in executing the generated code.

For instance, when testing LLMs on the Bike Sharing competition, we encountered 131 failures ($68.95\%$ of total failures) due to ``not found in axis'' errors. Among these, 62 cases ($47.33\%$ of not-found-in-axis errors) explicitly mentioned \texttt{datetime} in their error message, as described in the following. When tasked with predicting outcomes for the 20th to the end of the month using training data from the first 19 days of the month, LLMs repeatedly failed to handle the temporal progression. Instead of recognizing years as points on a continuous timeline, they treated them as distinct categories to handle. This misunderstanding led to runtime exceptions such as:

\begin{tcolorbox}[
    breakable,
    enhanced jigsaw,
    pad at break*=1mm,
    colback=blue!10,
    colframe=blue!50!black,
    arc=4mm,
    title=Python Code Snippet: LabelEncoder Error Due to Unseen Test Labels (1),
]
\begin{lstlisting}
Traceback (most recent call last):
  File "temp_generated.py", line 56, in <module>
    'datetime': test['datetime'],
  File "frame.py", line 4102, in __getitem__
    indexer = self.columns.get_loc(key)
  File "base.py", line 3812, in get_loc
    raise KeyError(key) from err
KeyError: 'datetime'
\end{lstlisting}
\end{tcolorbox}
Another 2 failures due to the same reason from Bike Sharing competition:
\begin{tcolorbox}[
    breakable,
    enhanced jigsaw,
    pad at break*=1mm,
    colback=blue!10,
    colframe=blue!50!black,
    arc=4mm,
    title=Python Code Snippet: LabelEncoder Error Due to Unseen Test Labels (2),
]
\begin{lstlisting}
ValueError: y contains previously unseen labels: [11, 12, 13, 14, 15, 16, 17, 18, 19]
\end{lstlisting}
\end{tcolorbox}

Additionally, we also observed that LLMs struggled with date parsing for time-and-date related data, such as in the Rossmann Store competition, where 10 failures involved date-related errors:
\begin{tcolorbox}[
    breakable,
    enhanced jigsaw,
    pad at break*=1mm,
    colback=blue!10,
    colframe=blue!50!black,
    arc=4mm,
    title=Python Code Snippet: Datetime Parsing Error Due to Format Mismatch,
]
\begin{lstlisting}
ValueError: time data '1/1/2012 0:00' does not match format '%Y-%m-%d %H:%M:%S'
\end{lstlisting}
\end{tcolorbox}

\subsection{Feature Engineering Alignment}

In our Kaggle experiments, we observed that LLMs often fail to align the training and testing features after applying feature engineering suggested by the provided domain knowledge, leading to runtime errors when executing the generated code.

For example, in the Bike Sharing competition, 99 cases (52.11\% of total failures) involved the LLM following notebook instructions to drop the features \texttt{casual} and \texttt{registered} (as demonstrated in the figure below) from the training data, as these are known to be highly correlated with other variables.
\begin{tcolorbox}[
    enhanced jigsaw,
    pad at break*=1mm,
    colback=blue!10,
    colframe=blue!50!black,
    arc=4mm,
    title=Python Code Snippet: Mismatched Feature Set Between Train and Test,
]
\begin{lstlisting}[language=Python]
# Drop features for training
X = train.drop(['datetime', 'casual', 'registered', 'count'], axis=1)
y = train['count']

# Drop features for test (incomplete)
X_test = test.drop(['datetime'], axis=1)
\end{lstlisting}
\end{tcolorbox}

Although this step is valid and important during training, the model applied it only to the training data and failed to perform the same transformation on the test data. As a result, the misaligned features between the training and test sets led to the following error:
\begin{tcolorbox}[
    breakable,
    enhanced jigsaw,
    pad at break*=1mm,
    colback=blue!10,
    colframe=blue!50!black,
    arc=4mm,
    title=Python Code Snippet: Feature Names Mismatch Between Training and Inference,
]
\begin{lstlisting}
ValueError: The feature names should match those that were passed during fit. 
Feature names unseen at fit time: casual, registered
\end{lstlisting}
\end{tcolorbox}

\subsection{Non-Numerical Data Handling Issues}

In our Kaggle experiments, we observed that LLMs have difficulty handling non-numerical data across all competitions.

For instance, type conversion errors accounted for 16 failures (7.11\%) in Bike Sharing, 78 failures (33.19\%) in BNP Paribas, 34 failures (13.03\%) in Rossmann, 16 failures (8.94\%) in Otto Group, and 19 failures (11.80\%) in Allstate. The most common pattern involved attempting to convert categorical values to numeric types:
\begin{tcolorbox}[
    breakable,
    enhanced jigsaw,
    pad at break*=1mm,
    colback=blue!10,
    colframe=blue!50!black,
    arc=4mm,
    title=Python Code Snippet: Scaling Error Due to Non-Numeric Feature Values,
]
\begin{lstlisting}
Traceback (most recent call last):
  File "generated.py", line 46, in <module>
    X_scaled = scaler.fit_transform(X)
...
ValueError: could not convert string to float: 'a'
\end{lstlisting}
\end{tcolorbox}

In BNP Paribas, 61 failures (25.96\%) showed similar errors with financial categorical data:

\begin{tcolorbox}[
    breakable,
    enhanced jigsaw,
    pad at break*=1mm,
    colback=blue!10,
    colframe=blue!50!black,
    arc=4mm,
    title=Python Code Snippet: Imputation Error Due to Non-Numeric Data with Mean Strategy,
]
\begin{lstlisting}
ValueError: Cannot use mean strategy with non-numeric data:
could not convert string to float: 'C'
\end{lstlisting}
\end{tcolorbox}

The Rossmann competition demonstrated particularly severe problems with categorical data aggregation, with 100 failures (38.31\%) showing:

\begin{tcolorbox}[
    breakable,
    enhanced jigsaw,
    pad at break*=1mm,
    colback=blue!10,
    colframe=blue!50!black,
    arc=4mm,
    title=Python Code Snippet: Aggregation Error Due to Non-Numeric Data in GroupBy Mean,
]
\begin{lstlisting}
Traceback (most recent call last):
  File "pandas/core/groupby/groupby.py", line 1946, in _agg_py_fallback
    raise type(err)(msg) from err
TypeError: agg function failed [how->mean,dtype->object]
\end{lstlisting}
\end{tcolorbox}

Feature name mismatches constituted 121 failures (53.78\%) in Bike Sharing due to inconsistent pre-processing:

\begin{tcolorbox}[
    breakable,
    enhanced jigsaw,
    pad at break*=1mm,
    colback=blue!10,
    colframe=blue!50!black,
    arc=4mm,
    title=Python Code Snippet: Prediction Error Due to Unseen Features at Inference,
]
\begin{lstlisting}
ValueError: The feature names should match those that were passed during fit.
Feature names unseen at fit time:
- casual
- registered
\end{lstlisting}
\end{tcolorbox}

In the Otto Group, 46 dimension errors (25.70\%) resulted from inappropriate handling of multi-dimensional categorical arrays:

\begin{tcolorbox}[
    breakable,
    enhanced jigsaw,
    pad at break*=1mm,
    colback=blue!10,
    colframe=blue!50!black,
    arc=4mm,
    title=Python Code Snippet: Dimensionality Error Due to Unexpected 3D Input Array,
]
\begin{lstlisting}
ValueError: Found array with dim 3. None expected <= 2.
\end{lstlisting}
\end{tcolorbox}

\section{Impact Statement}

This paper contributes to the field of Machine Learning by introducing a benchmark for evaluating LLM agents in practical data science scenarios. By systematically providing synthetic and Kaggle datasets paired with helpful and adversarial hints, our benchmark advances the evaluation of LLM capabilities in realistic and complex settings. Potential social benefits include improved transparency and reliability in automated data science processes, enhancing both AI interpretability and practical decision-making. While our work does not directly involve human subjects or sensitive data, the automation of data science workflows using LLMs could have broader societal implications, such as impacting employment in certain sectors. We encourage future research to explore these ethical considerations in more depth. 

\section{Use of AI Assistants}

We used ChatGPT to improve the writing when preparing this manuscript. All content generated or revised with this AI assistant was carefully reviewed and edited by us. We affirm that we take full responsibility for the final content presented in this work.

\section{Detailed Table for Experimental Results}\label{appendix:results}

\subsection{Results for Synthetic Datasets}

Table~\ref{tab:diabetes} to \ref{tab:wine_quality} present experimental results on performance scores and performance changes in each domain of synthetic datasets. Table~\ref{tab:syntask1_o4} to \ref{tab:syntask5_o4} present experimental results for GPT-o4-mini, GPT-4o, and GPT-4o-mini on submission rates, selection rates, and performance scores per task. All single experiments with LLMs are replicated for 5 times. There is no temperature control for GPT-o4-mini, and we set its reasoning effort as "medium". The rest of the LLMs are using $\text{temperature} = 0.2$. 

\begin{table*}[htbp]
\centering
\caption{Diabetes Performance Results. \textbf{Macro-F1} (higher is better). Format: value (\%change from baseline).}
\label{tab:diabetes}
\scalebox{0.85}{
\begin{tabular}{lccccc}
\toprule
\textbf{Model} & \textbf{None} & \textbf{Helpful-only} & \textbf{Adversarial-only} & \textbf{Mixed} & \textbf{Mixed (Misleading)} \\
\midrule
GPT-o4-mini & 66.33 & 69.32 (+4.50\%) & 62.19 (-6.25\%) & 68.25 (+2.90\%) & 55.08 (-16.96\%) \\
GPT-4o & 67.87 & 69.49 (+2.39\%) & 61.90 (-8.79\%) & 61.67 (-9.13\%) & 56.95 (-16.09\%) \\
GPT-4o-mini & 64.61 & 69.47 (+7.51\%) & 54.76 (-15.25\%) & 59.03 (-8.65\%) & 56.40 (-12.71\%) \\
Claude 3.5 haiku & 66.46 & 69.87 (+5.14\%) & 65.59 (-1.31\%) & 70.12 (+5.51\%) & 63.22 (-4.87\%) \\
Gemini 2.0 flash & 63.66 & 69.42 (+9.05\%) & 61.22 (-3.83\%) & 65.01 (+2.13\%) & 53.98 (-15.20\%) \\
DeepSeek-chat & 66.41 & 69.53 (+4.70\%) & 62.21 (-6.32\%) & 64.00 (-3.63\%) & 54.87 (-17.37\%) \\
\midrule
\textit{Human (expert)} & 66.60 & 71.00 (+6.61\%) & 66.60 (+0.00\%) & 71.00 (+6.61\%) & 71.00 (+6.61\%) \\
\bottomrule
\end{tabular}}
\end{table*}

\begin{table*}[htbp]
\centering
\caption{Game Revenue Performance Results. \textbf{RMSE} (lower is better, positive \% change indicates improvement). Format: value (\%change from baseline).}
\label{tab:game_revenue}
\scalebox{0.85}{\begin{tabular}{lccccc}
\toprule
\textbf{Model} & \textbf{None} & \textbf{Helpful-only} & \textbf{Adversarial-only} & \textbf{Mixed} & \textbf{Mixed (Misleading)} \\
\midrule
GPT-o4-mini & 30.09 & 24.32 (+19.19\%) & 52.14 (-73.29\%) & 24.32 (+19.19\%) & 64.65 (-114.86\%) \\
GPT-4o & 30.39 & 18.72 (+38.42\%) & 71.41 (-134.95\%) & 68.57 (-125.60\%) & 67.83 (-123.15\%) \\
GPT-4o-mini & 30.18 & 22.19 (+26.47\%) & 64.92 (-115.09\%) & 61.83 (-104.86\%) & 61.33 (-103.20\%) \\
Claude 3.5 haiku & 30.07 & 21.73 (+27.75\%) & 66.47 (-121.01\%) & 23.95 (+20.36\%) & 69.34 (-130.57\%) \\
Gemini 2.0 flash & 30.50 & 22.03 (+27.77\%) & 57.90 (-89.82\%) & 45.41 (-48.87\%) & 61.93 (-103.02\%) \\
DeepSeek-chat & 30.07 & 21.72 (+27.77\%) & 58.05 (-93.04\%) & 43.91 (-45.99\%) & 65.66 (-118.31\%) \\
\midrule
\textit{Human (expert)} & 30.09 & 18.07 (+39.95\%) & 30.09 (+0.00\%) & 18.07 (+39.95\%) & 18.07 (+39.95\%) \\
\bottomrule
\end{tabular}}
\end{table*}

\begin{table*}[htbp]
\centering
\caption{Haircut Rate Performance Results. \textbf{Macro-F1} (higher is better). Format: value (\%change from baseline).}
\label{tab:haircut_rate}
\scalebox{0.85}{\begin{tabular}{lccccc}
\toprule
\textbf{Model} & \textbf{None} & \textbf{Helpful-only} & \textbf{Adversarial-only} & \textbf{Mixed} & \textbf{Mixed (Misleading)} \\
\midrule
GPT-o4-mini & 17.67 & 18.38 (+4.00\%) & 15.08 (-14.70\%) & 24.49 (+38.56\%) & 12.67 (-28.33\%) \\
GPT-4o & 17.40 & 27.87 (+60.18\%) & 9.00 (-48.29\%) & 18.86 (+8.38\%) & 13.30 (-23.54\%) \\
GPT-4o-mini & 17.51 & 25.03 (+43.00\%) & 12.75 (-27.18\%) & 17.94 (+2.46\%) & 17.93 (+2.44\%) \\
Claude 3.5 haiku & 17.29 & 31.08 (+79.77\%) & 16.12 (-6.76\%) & 22.09 (+27.74\%) & 11.17 (-35.42\%) \\
Gemini 2.0 flash & 17.09 & 25.93 (+51.76\%) & 16.73 (-2.08\%) & 19.08 (+11.64\%) & 14.00 (-18.07\%) \\
DeepSeek-chat & - & 27.92 & 17.56 & 21.11 & 14.24 \\
\midrule
\textit{Human (expert)} & 17.50 & 27.30 (+56.00\%) & 17.50 (+0.00\%) & 27.30 (+56.00\%) & 27.30 (+56.00\%) \\
\bottomrule
\end{tabular}}
\end{table*}

\begin{table*}[htbp]
\centering
\caption{Housekeeping Performance Results. \textbf{Macro-F1} (higher is better). Format: value (\%change from baseline).}
\label{tab:housekeeping}
\scalebox{0.85}{\begin{tabular}{lccccc}
\toprule
\textbf{Model} & \textbf{None} & \textbf{Helpful-only} & \textbf{Adversarial-only} & \textbf{Mixed} & \textbf{Mixed (Misleading)} \\
\midrule
GPT-o4-mini & 59.17 & 65.63 (+10.93\%) & 56.05 (-5.28\%) & 66.61 (+12.58\%) & 52.59 (-11.12\%) \\
GPT-4o & 59.08 & 69.73 (+18.03\%) & 52.52 (-11.10\%) & 63.51 (+7.50\%) & 52.55 (-11.06\%) \\
GPT-4o-mini & 58.38 & 69.77 (+19.51\%) & 51.95 (-11.02\%) & 61.92 (+6.08\%) & 60.90 (+4.32\%) \\
Claude 3.5 haiku & 59.34 & 65.69 (+10.71\%) & 57.38 (-3.30\%) & 68.30 (+15.11\%) & 53.19 (-10.37\%) \\
Gemini 2.0 flash & 58.98 & 69.31 (+17.52\%) & 55.62 (-5.70\%) & 66.35 (+12.51\%) & 52.70 (-10.64\%) \\
DeepSeek-chat & 59.30 & - & 52.75 (-11.04\%) & - & 52.97 (-10.66\%) \\
\midrule
\textit{Human (expert)} & 59.50 & 73.43 (+23.41\%) & 59.50 (+0.00\%) & 73.43 (+23.41\%) & 73.43 (+23.41\%) \\
\bottomrule
\end{tabular}}
\end{table*}

\begin{table*}[htbp]
\centering
\caption{Machine Failure Performance Results. \textbf{Macro-F1} (higher is better). Format: value (\%change from baseline).}
\label{tab:machine_failure}
\scalebox{0.85}{\begin{tabular}{lccccc}
\toprule
\textbf{Model} & \textbf{None} & \textbf{Helpful-only} & \textbf{Adversarial-only} & \textbf{Mixed} & \textbf{Mixed (Misleading)} \\
\midrule
GPT-o4-mini & 75.47 & 78.74 (+4.34\%) & 73.48 (-2.63\%) & 80.19 (+6.26\%) & 65.30 (-13.48\%) \\
GPT-4o & 75.83 & 84.67 (+11.66\%) & 70.90 (-6.50\%) & 76.92 (+1.44\%) & 64.89 (-14.43\%) \\
GPT-4o-mini & 75.59 & 84.21 (+11.41\%) & 61.93 (-18.06\%) & 76.17 (+0.77\%) & 75.30 (-0.38\%) \\
Claude 3.5 haiku & 75.52 & 84.73 (+12.20\%) & 68.84 (-8.85\%) & 84.73 (+12.20\%) & 65.08 (-13.82\%) \\
Gemini 2.0 flash & 75.55 & 83.78 (+10.89\%) & 68.43 (-9.42\%) & 75.75 (+0.26\%) & 65.54 (-13.25\%) \\
DeepSeek-chat & 75.52 & 87.25 (+15.53\%) & 61.23 (-18.93\%) & 75.58 (+0.08\%) & 67.62 (-10.46\%) \\
\midrule
\textit{Human (expert)} & 75.61 & 88.80 (+17.46\%) & 75.61 (+0.00\%) & 88.80 (+17.46\%) & 88.80 (+17.46\%) \\
\bottomrule
\end{tabular}}
\end{table*}

\begin{table*}[htbp]
\centering
\caption{Power Generation Performance Results. \textbf{RMSE} (lower is better, positive \% change indicates improvement). Format: value (\%change from baseline).}
\label{tab:power_generation}
\scalebox{0.85}{\begin{tabular}{lccccc}
\toprule
\textbf{Model} & \textbf{None} & \textbf{Helpful-only} & \textbf{Adversarial-only} & \textbf{Mixed} & \textbf{Mixed (Misleading)} \\
\midrule
GPT-o4-mini & 228.22 & 230.04 (-0.80\%) & 229.37 (-0.50\%) & 229.29 (-0.47\%) & 230.37 (-0.94\%) \\
GPT-4o & 522.98 & 517.43 (+1.06\%) & 532.21 (-1.77\%) & 520.32 (+0.51\%) & 534.89 (-2.28\%) \\
GPT-4o-mini & 358.14 & - & 426.30 (-19.03\%) & 478.16 (-33.51\%) & 522.11 (-45.79\%) \\
Claude 3.5 haiku & 237.20 & 238.94 (-0.74\%) & 249.82 (-5.32\%) & 242.87 (-2.39\%) & 304.48 (-28.37\%) \\
Gemini 2.0 flash & 237.88 & 235.74 (+0.90\%) & 230.23 (+3.22\%) & 248.68 (-4.54\%) & 230.13 (+3.26\%) \\
DeepSeek-chat & 287.41 & 259.71 (+9.64\%) & 230.33 (+19.86\%) & 231.18 (+19.56\%) & 261.00 (+9.19\%) \\
\midrule
\textit{Human (expert)} & 521.64 & 501.50 (+3.86\%) & 521.64 (+0.00\%) & 501.50 (+3.86\%) & 501.50 (+3.86\%) \\
\bottomrule
\end{tabular}}
\end{table*}

\begin{table*}[htbp]
\centering
\caption{Real Estate Performance Results. \textbf{RMSE} (lower is better, positive \% change indicates improvement). Format: value (\%change from baseline).}
\label{tab:real_estate}
\scalebox{0.85}{\begin{tabular}{lccccc}
\toprule
\textbf{Model} & \textbf{None} & \textbf{Helpful-only} & \textbf{Adversarial-only} & \textbf{Mixed} & \textbf{Mixed (Misleading)} \\
\midrule
GPT-o4-mini & 44.15 & 38.05 (+13.82\%) & 55.70 (-26.17\%) & 31.70 (+28.19\%) & 61.36 (-38.98\%) \\
GPT-4o & 44.19 & 28.05 (+36.51\%) & 57.62 (-30.40\%) & 45.44 (-2.83\%) & 62.73 (-41.96\%) \\
GPT-4o-mini & 44.26 & 20.50 (+53.69\%) & 62.61 (-41.46\%) & 45.94 (-3.79\%) & 54.67 (-23.53\%) \\
Claude 3.5 haiku & 44.14 & 27.92 (+36.75\%) & 56.88 (-28.85\%) & 33.38 (+24.38\%) & 51.01 (-15.56\%) \\
Gemini 2.0 flash & 42.91 & 29.43 (+31.42\%) & 59.18 (-37.90\%) & 39.15 (+8.77\%) & 57.46 (-33.91\%) \\
DeepSeek-chat & 44.17 & 27.59 (+37.52\%) & 56.58 (-28.11\%) & 49.56 (-12.20\%) & 61.80 (-39.93\%) \\
\midrule
\textit{Human (expert)} & 44.15 & 20.22 (+54.21\%) & 44.15 (+0.00\%) & 20.22 (+54.21\%) & 20.22 (+54.21\%) \\
\bottomrule
\end{tabular}}
\end{table*}

\begin{table*}[htbp]
\centering
\caption{Second-hand Goods Performance Results. \textbf{RMSE} (lower is better, positive \% change indicates improvement). Format: value (\%change from baseline).}
\label{tab:second_hand_goods}
\scalebox{0.85}{\begin{tabular}{lccccc}
\toprule
\textbf{Model} & \textbf{None} & \textbf{Helpful-only} & \textbf{Adversarial-only} & \textbf{Mixed} & \textbf{Mixed (Misleading)} \\
\midrule
GPT-o4-mini & 3.44 & 3.41 (+0.71\%) & 3.52 (-2.34\%) & 3.37 (+1.96\%) & 3.59 (-4.53\%) \\
GPT-4o & 3.38 & 3.32 (+1.70\%) & 3.59 (-6.22\%) & 3.50 (-3.62\%) & 3.57 (-5.80\%) \\
GPT-4o-mini & 3.37 & 3.30 (+1.96\%) & 3.59 (-6.49\%) & 3.49 (-3.57\%) & 3.47 (-2.98\%) \\
Claude 3.5 haiku & 3.39 & 3.29 (+2.77\%) & 3.49 (-3.00\%) & 3.31 (+2.38\%) & 3.55 (-4.92\%) \\
Gemini 2.0 flash & 3.74 & 3.34 (+10.59\%) & 3.47 (+7.09\%) & 3.42 (+8.43\%) & 3.61 (+3.50\%) \\
DeepSeek-chat & 3.37 & 3.25 (+3.64\%) & 3.58 (-6.11\%) & 3.54 (-5.14\%) & 3.57 (-6.02\%) \\
\midrule
\textit{Human (expert)} & 3.39 & 3.26 (+3.98\%) & 3.39 (+0.00\%) & 3.26 (+3.98\%) & 3.26 (+24.07\%) \\
\bottomrule
\end{tabular}}
\end{table*}

\begin{table*}[htbp]
\centering
\caption{Song Popularity Performance Results. \textbf{Macro-F1} (higher is better). Format: value (\%change from baseline).}
\label{tab:song_popularity}
\scalebox{0.85}{\begin{tabular}{lccccc}
\toprule
\textbf{Model} & \textbf{None} & \textbf{Helpful-only} & \textbf{Adversarial-only} & \textbf{Mixed} & \textbf{Mixed (Misleading)} \\
\midrule
GPT-o4-mini & 39.26 & 43.36 (+10.44\%) & 33.38 (-14.97\%) & 44.21 (+12.61\%) & 30.31 (-22.79\%) \\
GPT-4o & 40.64 & 47.79 (+17.60\%) & 30.31 (-25.43\%) & 38.39 (-5.54\%) & 30.36 (-25.29\%) \\
GPT-4o-mini & 42.44 & 48.71 (+14.76\%) & 30.59 (-27.93\%) & 39.30 (-7.41\%) & 38.07 (-10.30\%) \\
Claude 3.5 haiku & 39.77 & 47.45 (+19.33\%) & 31.48 (-20.84\%) & 44.42 (+11.71\%) & 38.57 (-3.02\%) \\
Gemini 2.0 flash & 39.55 & 48.61 (+22.91\%) & 30.29 (-23.43\%) & 41.87 (+5.87\%) & 30.55 (-22.76\%) \\
DeepSeek-chat & 43.33 & 50.41 (+16.34\%) & - & 47.90 (+10.54\%) & 32.63 (-24.70\%) \\
\midrule
\textit{Human (expert)} & 41.42 & 51.39 (+24.07\%) & 41.42 (+0.00\%) & 51.39 (+24.07\%) & 51.39 (+24.07\%) \\
\bottomrule
\end{tabular}}
\end{table*}

\begin{table*}[htbp]
\centering
\caption{Wine Quality Performance Results. \textbf{Macro-F1} (higher is better). Format: value (\%change from baseline).}
\label{tab:wine_quality}
\scalebox{0.85}{\begin{tabular}{lccccc}
\toprule
\textbf{Model} & \textbf{None} & \textbf{Helpful-only} & \textbf{Adversarial-only} & \textbf{Mixed} & \textbf{Mixed (Misleading)} \\
\midrule
GPT-o4-mini & 28.72 & 33.86 (+17.91\%) & 19.84 (-30.93\%) & 25.12 (-12.53\%) & 19.80 (-31.06\%) \\
GPT-4o & 27.59 & 33.16 (+20.21\%) & 19.90 (-27.87\%) & 24.63 (-10.73\%) & 20.90 (-24.23\%) \\
GPT-4o-mini & 27.44 & 33.06 (+20.50\%) & 20.17 (-26.50\%) & 24.46 (-10.85\%) & 24.48 (-10.77\%) \\
Claude 3.5 haiku & 28.43 & 33.90 (+19.25\%) & 19.71 (-30.69\%) & 27.59 (-2.95\%) & 22.96 (-19.23\%) \\
Gemini 2.0 flash & 28.01 & 33.83 (+20.77\%) & 22.29 (-20.44\%) & 24.55 (-12.37\%) & 20.71 (-26.06\%) \\
DeepSeek-chat & 29.03 & 34.37 (+18.38\%) & 20.86 (-28.13\%) & 27.57 (-5.02\%) & 20.05 (-30.94\%) \\
\midrule
\textit{Human (expert)} & 28.42 & 35.01 (+23.20\%) & 28.42 (+0.00\%) & 35.01 (+23.20\%) & 35.01 (+23.20\%) \\
\bottomrule
\end{tabular}}
\end{table*}

\begin{table*}[htbp]
\centering
\caption{Synthetic dataset results with GPT-o4-mini given helpful documents only. For classification domains, Macro-F1 \% (mean $\pm$ std); for regression domains, RMSE (mean $\pm$ std). Selection rates in \%.}\label{tab:syntask1_o4}
\scalebox{0.9}{
\begin{tabular}{llccc}
\toprule
\textbf{Domain} & \textbf{Metric} & \multicolumn{3}{c}{\textbf{No. of Helpful Hints}} \\
\cmidrule(lr){3-5}
 &  & \textbf{0} & \textbf{1} & \textbf{2} \\
\midrule
\multirow{4}{*}{Diabetes}
 & Valid submission \%      & 100 & 40 & 100 \\
 & Macro-F1 \%              & $66.33\pm0.21$ & $67.28\pm1.55$ & $70.13\pm2.11$ \\
 & Helpful selection \%     & - & $100\pm0.00$ & $100\pm0.00$ \\
\midrule
\multirow{4}{*}{Haircut Rate}
 & Valid submission \%      & 100 & 60 & 100 \\
 & Macro-F1 \%              & $17.67\pm0.21$ & $22.45\pm0.00$ & $17.57\pm0.15$ \\
 & Helpful selection \%     & - & $100\pm0.00$ & $100\pm0.00$ \\
\bottomrule
\end{tabular}
}
\end{table*}

\begin{table*}[htbp]
\centering
\caption{Synthetic dataset results with GPT-4o given helpful documents only. For classification domains, Macro-F1 \% (mean $\pm$ std); for regression domains, RMSE (mean $\pm$ std). Selection rates in \%.}\label{tab:syntask1_4o}
\scalebox{0.9}{
\begin{tabular}{llccc}
\toprule
\textbf{Domain} & \textbf{Metric} & \multicolumn{3}{c}{\textbf{No. of Helpful Hints}} \\
\cmidrule(lr){3-5}
 &  & \textbf{0} & \textbf{1} & \textbf{2} \\
\midrule
\multirow{4}{*}{Diabetes}
 & Valid submission \%      & 100 & 100 & 100 \\
 & Macro-F1 \%              & $67.87\pm1.86$ & $68.09\pm0.13$ & $70.89\pm0.18$ \\
 & Helpful selection \%     & - & $100\pm0.00$ & $100\pm0.00$ \\
\midrule
\multirow{4}{*}{Haircut Rate}
 & Valid submission \%      & 40 & 0 & 60 \\
 & Macro-F1 \%              & $17.40\pm0.00$ & – & $27.87\pm0.60$ \\
 & Helpful selection \%     & - & $100\pm0.00$ & $100\pm0.00$ \\
\bottomrule
\end{tabular}
}
\end{table*}

\begin{table*}[htbp]
\centering
\caption{Synthetic dataset results with GPT-4o-mini given helpful documents only. For classification domains, Macro-F1 \% (mean $\pm$ std); for regression domains, RMSE (mean $\pm$ std). Selection rates in \%.}\label{tab:syntask1_4omini}
\scalebox{0.9}{
\begin{tabular}{llccc}
\toprule
\textbf{Domain} & \textbf{Metric} & \multicolumn{3}{c}{\textbf{No. of Helpful Hints}} \\
\cmidrule(lr){3-5}
 &  & \textbf{0} & \textbf{1} & \textbf{2} \\
\midrule
\multirow{4}{*}{Diabetes}
 & Valid submission \%      & 100 & 100 & 100 \\
 & Macro-F1 \%              & $64.61\pm2.41$ & $68.03\pm0.04$ & $70.91\pm0.16$ \\
 & Helpful selection \%     & - & $100\pm0.00$ & $100\pm0.00$ \\
\midrule
\multirow{4}{*}{Haircut Rate}
 & Valid submission \%      & 100 & 100 & 100 \\
 & Macro-F1 \%              & $17.51\pm0.14$ & $22.55\pm0.00$ & $27.52\pm0.00$ \\
 & Helpful selection \%     & - & $100\pm0.00$ & $100\pm0.00$ \\
\bottomrule
\end{tabular}
}
\end{table*}

\begin{table*}[htbp]
\centering
\caption{Synthetic dataset results with GPT-o4-mini given only adversarial documents. For classification domains, Macro-F1 \% (mean $\pm$ std); for regression domains, RMSE (mean $\pm$ std). Selection rates in \%.}\label{tab:syntask2_o4}
\scalebox{0.9}{
\begin{tabular}{llccc}
\toprule
\textbf{Domain} & \textbf{Metric} & \multicolumn{3}{c}{\textbf{No. of Adversarial Documents}} \\
\cmidrule(lr){3-5}
 &  & \textbf{0} & \textbf{1} & \textbf{2} \\
\midrule
\multirow{4}{*}{Diabetes}
 & Valid submission \%      & 100 & 100 & 100 \\
 & Macro-F1 \%              & $66.33\pm0.21$ & $57.97\pm0.12$ & $66.41\pm0.21$ \\
 & Adversarial selection \% & - & $100\pm0.00$ & $0.00\pm0.00$ \\
\midrule
\multirow{4}{*}{Haircut Rate}
 & Valid submission \%      & 100 & 100 & 80 \\
 & Macro-F1 \%              & $17.67\pm0.21$ & $16.77\pm0.13$ & $13.38\pm4.82$ \\
 & Adversarial selection \% & - & $100\pm0.00$ & $40.00\pm54.77$ \\
\bottomrule
\end{tabular}
}
\end{table*}

\begin{table*}[htbp]
\centering
\caption{Synthetic dataset results with GPT-4o given only adversarial documents. For classification domains, Macro-F1 \% (mean $\pm$ std); for regression domains, RMSE (mean $\pm$ std). Selection rates in \%.}\label{tab:syntask2_4o}
\scalebox{0.9}{
\begin{tabular}{llccc}
\toprule
\textbf{Domain} & \textbf{Metric} & \multicolumn{3}{c}{\textbf{No. of Adversarial Documents}} \\
\cmidrule(lr){3-5}
 &  & \textbf{0} & \textbf{1} & \textbf{2} \\
\midrule
\multirow{4}{*}{Diabetes}
 & Valid submission \%      & 100 & 80 & 100 \\
 & Macro-F1 \%              & $67.87\pm1.86$ & $59.75\pm2.73$ & $63.62\pm3.16$ \\
 & Adversarial selection \% & - & $100\pm0.00$ & $40.00\pm54.77$ \\
\midrule
\multirow{4}{*}{Haircut Rate}
 & Valid submission \%      & 40 & 0 & 40 \\
 & Macro-F1 \%              & $17.40\pm0.00$ & – & $9.00\pm0.00$ \\
 & Adversarial selection \% & - & $100\pm0.00$ & $100\pm0.00$ \\
\bottomrule
\end{tabular}
}
\end{table*}

\begin{table*}[htbp]
\centering
\caption{Synthetic dataset results with GPT-4o-mini given only adversarial documents. For classification domains, Macro-F1 \% (mean $\pm$ std); for regression domains, RMSE (mean $\pm$ std). Selection rates in \%.}\label{tab:syntask2_4omini}
\scalebox{0.9}{
\begin{tabular}{llccc}
\toprule
\textbf{Domain} & \textbf{Metric} & \multicolumn{3}{c}{\textbf{No. of Adversarial Documents}} \\
\cmidrule(lr){3-5}
 &  & \textbf{0} & \textbf{1} & \textbf{2} \\
\midrule
\multirow{4}{*}{Diabetes}
 & Valid submission \%      & 100 & 100 & 100 \\
 & Macro-F1 \%              & $64.61\pm2.41$ & $57.16\pm1.47$ & $52.36\pm0.28$ \\
 & Adversarial selection \% & - & $100\pm0.00$ & $100\pm0.00$ \\
\midrule
\multirow{4}{*}{Haircut Rate}
 & Valid submission \%      & 100 & 100 & 100 \\
 & Macro-F1 \%              & $17.51\pm0.14$ & $16.59\pm0.00$ & $8.90\pm0.13$ \\
 & Adversarial selection \% & - & $100\pm0.00$ & $100\pm0.00$ \\
\bottomrule
\end{tabular}
}
\end{table*}

\begin{table*}[htbp]
\centering
\caption{Synthetic dataset results with GPT-o4-mini given mixed helpful and adversarial documents. For classification domains, Macro-F1 \% (mean $\pm$ std); for regression domains, RMSE (mean $\pm$ std). Selection rates in \%.}\label{tab:syntask3_o4}
\scalebox{0.85}{
\begin{tabular}{llcccc}
\toprule
\textbf{Domain} & \textbf{Metric} & \multicolumn{4}{c}{\textbf{Helpful:Adversarial Ratio}} \\
\cmidrule(lr){3-6}
 &  & \textbf{1:1} & \textbf{1:2} & \textbf{2:1} & \textbf{2:2} \\
\midrule
\multirow{5}{*}{Diabetes}
 & Valid submission \%      & 100 & 100 & 100 & 80 \\
 & Macro-F1 \%              & $66.84\pm1.41$ & $67.55\pm0.90$ & $70.10\pm2.10$ & $68.59\pm2.56$ \\
 & Helpful selection \%     & $100\pm0.00$ & $100\pm0.00$ & $100\pm0.00$ & $100\pm0.00$ \\
 & Adversarial selection \% & $100\pm0.00$ & $40.00\pm54.77$ & $100\pm0.00$ & $60.00\pm54.77$ \\
\midrule
\multirow{5}{*}{Haircut Rate}
 & Valid submission \%      & 100 & 60 & 80 & 60 \\
 & Macro-F1 \%              & $19.64\pm2.92$ & $22.28\pm0.18$ & $30.75\pm1.79$ & $26.90\pm8.34$ \\
\bottomrule
\end{tabular}
}
\end{table*}

\begin{table*}[htbp]
\centering
\caption{Synthetic dataset results with GPT 4o given mixed helpful and adversarial documents. For classification domains, Macro-F1 \% (mean $\pm$ std); for regression domains, RMSE (mean $\pm$ std). Selection rates in \%.}\label{tab:syntask3_4o}
\scalebox{0.85}{
\begin{tabular}{llcccc}
\toprule
\textbf{Domain} & \textbf{Metric} & \multicolumn{4}{c}{\textbf{Helpful:Adversarial Ratio}} \\
\cmidrule(lr){3-6}
 &  & \textbf{1:1} & \textbf{1:2} & \textbf{2:1} & \textbf{2:2} \\
\midrule
\multirow{5}{*}{Diabetes}
 & Valid submission \%      & 100 & 100 & 100 & 60 \\
 & Macro-F1 \%              & $60.38\pm0.26$ & $60.31\pm5.39$ & $66.54\pm0.09$ & $57.97\pm0.48$ \\
 & Helpful selection \%     & $100\pm0.00$ & $100\pm0.00$ & $100\pm0.00$ & $100\pm0.00$ \\
 & Adversarial selection \% & $100\pm0.00$ & $70.00\pm27.39$ & $100\pm0.00$ & $100\pm0.00$ \\
\midrule
\multirow{5}{*}{Haircut Rate}
 & Valid submission \%      & 80 & 60 & 100 & 80 \\
 & Macro-F1 \%              & $20.08\pm0.54$ & $11.34\pm0.00$ & $26.29\pm0.27$ & $13.97\pm0.00$ \\
\bottomrule
\end{tabular}
}
\end{table*}

\begin{table*}[htbp]
\centering
\caption{Synthetic dataset results with GPT 4o-mini given mixed helpful and adversarial documents. For classification domains, Macro-F1 \% (mean $\pm$ std); for regression domains, RMSE (mean $\pm$ std). Selection rates in \%.}\label{tab:syntask3_4omini}
\scalebox{0.85}{
\begin{tabular}{llcccc}
\toprule
\textbf{Domain} & \textbf{Metric} & \multicolumn{4}{c}{\textbf{Helpful:Adversarial Ratio}} \\
\cmidrule(lr){3-6}
 &  & \textbf{1:1} & \textbf{1:2} & \textbf{2:1} & \textbf{2:2} \\
\midrule
\multirow{5}{*}{Diabetes}
 & Valid submission \%      & 100 & 100 & 100 & 80 \\
 & Macro-F1 \%              & $60.51\pm0.32$ & $51.14\pm0.00$ & $65.93\pm0.84$ & $58.41\pm0.00$ \\
 & Helpful selection \%     & $100\pm0.00$ & $100\pm0.00$ & $100\pm0.00$ & $100\pm0.00$ \\
 & Adversarial selection \% & $100\pm0.00$ & $100\pm0.00$ & $100\pm0.00$ & $100\pm0.00$ \\
\midrule
\multirow{5}{*}{Haircut Rate}
 & Valid submission \%      & 100 & 100 & 100 & 100 \\
 & Macro-F1 \%              & $20.35\pm0.00$ & $11.03\pm0.50$ & $26.39\pm0.00$ & $13.97\pm0.00$ \\
\bottomrule
\end{tabular}
}
\end{table*}

\begin{table*}[htbp]
\centering
\caption{Synthetic dataset results with GPT o4‑mini given mixed helpful and adversarial documents with misleading prompt. For classification domains, Macro‑F1 \% (mean $\pm$ std); for regression domains, RMSE (mean $\pm$ std). Selection rates in \%.}\label{tab:syntask5_o4}
\scalebox{0.85}{
\begin{tabular}{llcccc}
\toprule
\textbf{Domain} & \textbf{Metric} & \multicolumn{4}{c}{\textbf{Helpful:Adversarial Ratio}} \\
\cmidrule(lr){3-6}
 &  & \textbf{1:1} & \textbf{1:2} & \textbf{2:1} & \textbf{2:2} \\
\midrule
\multirow{4}{*}{Diabetes}
 & Valid submission \%      & 100 & 100 & 100 & 100 \\
 & Macro-F1 \%              & $58.06\pm0.19$ & $51.54\pm0.00$ & $59.18\pm2.69$ & $51.55\pm0.01$ \\
 & Helpful selection \%     & $0.00\pm0.00$ & $20.00\pm 40.00$ & $0.00\pm0.00$ & $0.00\pm0.00$ \\
 & Adversarial selection \% & $100\pm0.00$ & $100\pm0.00$ & $100\pm0.00$ & $100\pm0.00$ \\
\midrule
\multirow{4}{*}{Haircut Rate}
 & Valid submission \%      & 100 & 100 & 100 & 100 \\
 & Macro-F1 \%              & $16.91\pm0.00$ & $9.25\pm0.42$ & $16.93\pm0.07$ & $9.27\pm0.62$ \\
\bottomrule
\end{tabular}
}
\end{table*}

\begin{table*}[htbp]
\centering
\caption{Synthetic dataset results with GPT-4o given mixed helpful and adversarial documents with misleading prompt. For classification domains, Macro-F1 \% (mean $\pm$ std); for regression domains, RMSE (mean $\pm$ std). Selection rates in \%.}\label{tab:syntask5_4o}
\scalebox{0.85}{
\begin{tabular}{llcccc}
\toprule
\textbf{Domain} & \textbf{Metric} & \multicolumn{4}{c}{\textbf{Helpful:Adversarial Ratio}} \\
\cmidrule(lr){3-6}
 &  & \textbf{1:1} & \textbf{1:2} & \textbf{2:1} & \textbf{2:2} \\
\midrule
\multirow{5}{*}{Diabetes}
 & Valid submission \%      & 100 & 100 & 100 & 100 \\
 & Macro-F1 \%              & $60.48\pm4.73$ & $50.98\pm0.25$ & $59.54\pm4.16$ & $56.79\pm3.21$ \\
 & Helpful selection \%     & $0.00\pm0.00$ & $60.00\pm54.77$ & $0.00\pm0.00$ & $100\pm0.00$ \\
 & Adversarial selection \% & $100\pm0.00$ & $100\pm0.00$ & $100\pm0.00$ & $100\pm0.00$ \\
\midrule
\multirow{5}{*}{Haircut Rate}
 & Valid submission \%      & 20 & 80 & 80 & 40 \\
 & Macro-F1 \%              & $16.59\pm0.00$ & $11.34\pm0.00$ & $16.59\pm0.00$ & $9.00\pm0.00$ \\
\bottomrule
\end{tabular}
}
\end{table*}

\begin{table*}[htbp]
\centering
\caption{Synthetic dataset results with GPT 4o‑mini given mixed helpful and adversarial documents with misleading prompt. For classification domains, Macro‑F1 \% (mean $\pm$ std); for regression domains, RMSE (mean $\pm$ std). Selection rates in \%.}\label{tab:syntask5_4omini}
\scalebox{0.85}{
\begin{tabular}{llcccc}
\toprule
\textbf{Domain} & \textbf{Metric} & \multicolumn{4}{c}{\textbf{Helpful:Adversarial Ratio}} \\
\cmidrule(lr){3-6}
 &  & \textbf{1:1} & \textbf{1:2} & \textbf{2:1} & \textbf{2:2} \\
\midrule
\multirow{4}{*}{Diabetes}
 & Valid submission \%      & 100 & 100 & 20 & 0 \\
 & Macro-F1 \%              & $59.96\pm0.55$ & $51.39\pm0.58$ & $63.64\pm0.00$ & – \\
 & Helpful selection \%     & $100\pm0.00$ & $80.00\pm 40.00$ & $100\pm0.00$ & $100\pm0.00$ \\
 & Adversarial selection \% & $100\pm0.00$ & $100\pm0.00$ & $100\pm0.00$ & $100\pm0.00$ \\
\midrule
\multirow{4}{*}{Haircut Rate}
 & Valid submission \%      & 100 & 100 & 100 & 100 \\
 & Macro-F1 \%              & $20.35\pm0.00$ & $11.01\pm0.47$ & $26.39\pm0.00$ & $13.97\pm0.00$ \\
\bottomrule
\end{tabular}
}
\end{table*}

\subsection{Results for Kaggle Datasets}\label{appendix:kaggleresults}

Table~\ref{tab:allstate_claims_severity} to \ref{tab:rossmann_store_sales} present experimental results on performance scores (MAE for allstate-claims-severity, RMSLE for bike-sharing-demand, log loss for bnp-paribas-cardif-claims-management
and otto-group-product-classification-challenge, and RMSPE for rossmann-store-sales) and performance changes in each domain of Kaggle datasets. Table~\ref{tab:task6_gpt-4o-mini_temp0.7} to \ref{tab:task5_o4-mini_temp0.7} present experimental results for GPT-o4-mini, GPT-4o, and GPT-4o-mini on submission rates and selection rates. There is no temperature control for GPT-o4-mini, and we set its reasoning effort as "medium". The rest of the LLMs are using $\text{temperature} = 0.7$.

\begin{table*}[htbp]
\centering
\caption{allstate-claims-severity Performance Results. \textbf{MAE} (lower is better, positive \% change indicates improvement). Format: value (\%change from baseline).}
\label{tab:allstate_claims_severity}
\scalebox{0.85}{
\begin{tabular}{lccccc}
\toprule
\textbf{Model} & \textbf{None} & \textbf{Helpful-only} & \textbf{Adversarial-only} & \textbf{Mixed} & \textbf{Mixed (Misleading)} \\
\midrule
GPT-o4-mini & 1325.80 & 1292.60 (+2.50\%) & 1345.90 (-1.52\%) & 1315.02 (+0.81\%) & 1342.63 (-1.27\%) \\
GPT-4o & 1363.05 & 1346.47 (+1.22\%) & 1400.95 (-2.78\%) & 1395.11 (-2.35\%) & 1381.54 (-1.36\%) \\
GPT-4o-mini & — & 1334.75 & — & 1328.60 & 1388.53 \\
Claude 3.5 haiku & 1279.00 & 1340.28 (-4.79\%) & 1314.33 (-2.76\%) & 1372.98 (-7.35\%) & 1409.01 (-10.16\%) \\
Gemini 2.0 flash & 1220.45 & 1334.20 (-9.32\%) & 1471.40 (-20.56\%) & 1351.76 (-10.76\%) & 1504.55 (-23.28\%) \\
DeepSeek-chat & 1328.39 & 1325.59 (+0.21\%) & 1280.61 (+3.60\%) & 1362.16 (-2.54\%) & 1335.65 (-0.55\%) \\
\midrule
\textit{Human (expert)} & 1328.39 & 1257.35 (+5.65\%) & 1328.39 (+0.00\%) & 1257.35 (+5.65\%) & 1257.35 (+5.65\%) \\
\bottomrule
\end{tabular}}
\end{table*}

\begin{table*}[htbp]
\centering
\caption{bike-sharing-demand Performance Results. \textbf{RMSLE} (lower is better, positive \% change indicates improvement). Format: value (\%change from baseline).}
\label{tab:bike_sharing_demand}
\scalebox{0.85}{\begin{tabular}{lccccc}
\toprule
\textbf{Model} & \textbf{None} & \textbf{Helpful-only} & \textbf{Adversarial-only} & \textbf{Mixed} & \textbf{Mixed (Misleading)} \\
\midrule
GPT-o4-mini & 0.36 & 0.42 (-16.34\%) & 0.62 (-70.39\%) & 0.50 (-37.62\%) & 0.77 (-111.46\%) \\
GPT-4o & 0.94 & 0.57 (+39.99\%) & 0.39 (+58.83\%) & 0.38 (+59.84\%) & 0.77 (+18.02\%) \\
GPT-4o-mini & 1.40 & 0.36 (+73.99\%) & 0.36 (+74.05\%) & 0.36 (+73.91\%) & 0.36 (+74.05\%) \\
Claude 3.5 haiku & 0.36 & 0.36 (-0.02\%) & 0.36 (+0.35\%) & 0.36 (+0.18\%) & 0.36 (-0.07\%) \\
Gemini 2.0 flash & 0.40 & 0.36 (+9.92\%) & 0.37 (+6.18\%) & 0.37 (+7.58\%) & 0.66 (-66.13\%) \\
DeepSeek-chat & 0.77 & 0.36 (+52.91\%) & 0.36 (+52.83\%) & 0.36 (+52.81\%) & 0.75 (+2.18\%) \\
\midrule
\textit{Human (expert)} & 0.36 & 0.32 (+11.11\%) & 0.36 (+0.00\%) & 0.32 (+11.11\%) & 0.32 (+11.11\%) \\
\bottomrule
\end{tabular}}
\end{table*}

\begin{table*}[htbp]
\centering
\caption{bnp-paribas-cardif-claims-management Performance Results. \textbf{log loss} (lower is better, positive \% change indicates improvement). Format: value (\%change from baseline).}
\label{tab:bnp_paribas_cardif_claims_management}
\scalebox{0.85}{\begin{tabular}{lccccc}
\toprule
\textbf{Model} & \textbf{None} & \textbf{Helpful-only} & \textbf{Adversarial-only} & \textbf{Mixed} & \textbf{Mixed (Misleading)} \\
\midrule
GPT-o4-mini & 0.54 & 0.51 (+5.82\%) & 0.51 (+4.37\%) & 0.52 (+3.90\%) & 0.50 (+7.05\%) \\
GPT-4o & 0.50 & 0.50 (-0.52\%) & 0.50 (+0.91\%) & 0.50 (+0.06\%) & 0.51 (-1.64\%) \\
GPT-4o-mini & 0.50 & 0.50 (-0.38\%) & — & 0.51 (-2.60\%) & 0.50 (-0.45\%) \\
Claude 3.5 haiku & — & — & — & 0.51 & — \\
Gemini 2.0 flash & 0.60 & 0.50 (+17.50\%) & 0.50 (+17.98\%) & 0.51 (+15.66\%) & 0.52 (+13.67\%) \\
DeepSeek-chat & 0.50 & 0.51 (-1.43\%) & 0.51 (-2.41\%) & 0.52 (-4.03\%) & 0.53 (-5.81\%) \\
\midrule
\textit{Human (expert)} & 0.53 & 0.52 (+2.85\%) & 0.53 (+0.00\%) & 0.52 (+2.85\%) & 0.52 (+2.85\%) \\
\bottomrule
\end{tabular}}
\end{table*}

\begin{table*}[htbp]
\centering
\caption{otto-group-product-classification-challenge Performance Results. \textbf{log loss} (lower is better, positive \% change indicates improvement). Format: value (\%change from baseline).}
\label{tab:otto_group_product_classification_challenge}
\scalebox{0.85}{\begin{tabular}{lccccc}
\toprule
\textbf{Model} & \textbf{None} & \textbf{Helpful-only} & \textbf{Adversarial-only} & \textbf{Mixed} & \textbf{Mixed (Misleading)} \\
\midrule
GPT-o4-mini & 0.78 & 0.71 (+9.79\%) & 1.35 (-72.86\%) & 0.75 (+4.40\%) & 1.29 (-64.85\%) \\
GPT-4o & 0.77 & 0.59 (+23.61\%) & — & — & 2.69 (-249.52\%) \\
GPT-4o-mini & 0.93 & — & 0.79 (+15.05\%) & 2.39 (-156.99\%) & 2.41 (-159.14\%) \\
Claude 3.5 haiku & 0.78 & 1.87 (-141.04\%) & — & 0.82 (-5.36\%) & 0.99 (-26.95\%) \\
Gemini 2.0 flash & 0.78 & 0.78 (+0.17\%) & 1.36 (-74.48\%) & 1.75 (-125.02\%) & 1.62 (-107.70\%) \\
DeepSeek-chat & 0.78 & — & — & — & 1.42 (-81.54\%) \\
\midrule
\textit{Human (expert)} & 0.87 & 0.57 (+34.88\%) & 0.87 (+0.00\%) & 0.57 (+34.88\%) & 0.57 (+34.88\%) \\
\bottomrule
\end{tabular}}
\end{table*}

\begin{table*}[htbp]
\centering
\caption{rossmann-store-sales Performance Results. \textbf{RMSPE} (lower is better, positive \% change indicates improvement). Format: value (\%change from baseline).}
\label{tab:rossmann_store_sales}
\scalebox{0.85}{\begin{tabular}{lccccc}
\toprule
\textbf{Model} & \textbf{None} & \textbf{Helpful-only} & \textbf{Adversarial-only} & \textbf{Mixed} & \textbf{Mixed (Misleading)} \\
\midrule
GPT-o4-mini & 43.11 & 42.70 (+0.95\%) & 44.81 (-3.96\%) & 43.95 (-1.95\%) & 46.21 (-7.20\%) \\
GPT-4o & 42.00 & 47.74 (-13.66\%) & 47.80 (-13.80\%) & 46.26 (-10.13\%) & 46.31 (-10.27\%) \\
GPT-4o-mini & 40.33 & 42.99 (-6.59\%) & 47.08 (-16.72\%) & 43.17 (-7.04\%) & 45.74 (-13.41\%) \\
Claude 3.5 haiku & 40.34 & 40.37 (-0.09\%) & 46.06 (-14.20\%) & 45.90 (-13.79\%) & 48.01 (-19.02\%) \\
Gemini 2.0 flash & 44.33 & 43.57 (+1.71\%) & — & 67.97 (-53.34\%) & — \\
DeepSeek-chat & 41.32 & 42.33 (-2.44\%) & 50.76 (-22.85\%) & 46.47 (-12.48\%) & 51.14 (-23.76\%) \\
\midrule
\textit{Human (expert)} & 42.77 & 40.44 (+5.45\%) & 42.77 (+0.00\%) & 40.44 (+5.45\%) & 40.44 (+5.45\%) \\
\bottomrule
\end{tabular}}
\end{table*}

\begin{table*}[!htb]
\centering
\caption{Selection rates of adversarial notebooks by GPT-4o-mini across different datasets with varying helpful to adversarial notebook ratios. The table shows consistent preference for adversarial notebooks (nearly 100\% selection rate) regardless of dataset or notebook ratio, with minimal selection of helpful notebooks.}
\scalebox{0.9}{
\begin{tabular}{ccccc}
\toprule
\textbf{Dataset} & \textbf{Metric} & \multicolumn{3}{c}{\textbf{Helpful : Adversarial}} \\ \cline{3-5}
 &  & \textbf{1:3} & \textbf{2:3} & \textbf{3:1} \\
\midrule
\multirow{3}{*}{allstate-claims-severity}
 & Helpful selection\% & 0.0 $\pm$ 0.0 & 0.0 $\pm$ 0.0 & 0.0 $\pm$ 0.0 \\
 & Adversarial selection\% & 100.0 $\pm$ 0.0 & 100.0 $\pm$ 0.0 & 100.0 $\pm$ 0.0 \\
 & Valid submission\% & 100.0 & 80.0 & 80.0 \\
\midrule
\multirow{3}{*}{bike-sharing-demand}
 & Helpful selection\% & 0.0 $\pm$ 0.0 & 0.0 $\pm$ 0.0 & 6.7 $\pm$ 13.3 \\
 & Adversarial selection\% & 100.0 $\pm$ 0.0 & 100.0 $\pm$ 0.0 & 100.0 $\pm$ 0.0 \\
 & Valid submission\% & 40.0 & 0.0 & 40.0 \\
\midrule
\multirow{3}{*}{bnp-paribas-cardif-claims-management}
 & Helpful selection\% & 0.0 $\pm$ 0.0 & 0.0 $\pm$ 0.0 & 0.0 $\pm$ 0.0 \\
 & Adversarial selection\% & 100.0 $\pm$ 0.0 & 100.0 $\pm$ 0.0 & 100.0 $\pm$ 0.0 \\
 & Valid submission\% & 40.0 & 20.0 & 0.0 \\
\midrule
\multirow{3}{*}{otto-group-product-classification-challenge}
 & Helpful selection\% & 0.0 $\pm$ 0.0 & 0.0 $\pm$ 0.0 & 6.7 $\pm$ 13.3 \\
 & Adversarial selection\% & 100.0 $\pm$ 0.0 & 93.3 $\pm$ 13.3 & 100.0 $\pm$ 0.0 \\
 & Valid submission\% & 0.0 & 0.0 & 0.0 \\
\midrule
\multirow{3}{*}{rossmann-store-sales}
 & Helpful selection\% & 0.0 $\pm$ 0.0 & 0.0 $\pm$ 0.0 & 0.0 $\pm$ 0.0 \\
 & Adversarial selection\% & 100.0 $\pm$ 0.0 & 100.0 $\pm$ 0.0 & 100.0 $\pm$ 0.0 \\
 & Valid submission\% & 0.0 & 0.0 & 0.0 \\
\bottomrule
\end{tabular}
}
\label{tab:task6_gpt-4o-mini_temp0.7}
\end{table*}

\begin{table*}[!htb]
\centering
\caption{Selection rates of adversarial notebooks by GPT-4o across different datasets with varying helpful to adversarial notebook ratios. The table shows consistent preference for adversarial notebooks (nearly 100\% selection rate) regardless of dataset or notebook ratio, with minimal selection of helpful notebooks.}
\scalebox{0.9}{
\begin{tabular}{ccccc}
\toprule
\textbf{Dataset} & \textbf{Metric} & \multicolumn{3}{c}{\textbf{Helpful : Adversarial}} \\ \cline{3-5}
 &  & \textbf{1:3} & \textbf{2:3} & \textbf{3:1} \\
\midrule
\multirow{3}{*}{allstate-claims-severity}
 & Helpful selection\% & 0.0 $\pm$ 0.0 & 0.0 $\pm$ 0.0 & 0.0 $\pm$ 0.0 \\
 & Adversarial selection\% & 93.3 $\pm$ 13.3 & 100.0 $\pm$ 0.0 & 100.0 $\pm$ 0.0 \\
 & Valid submission\% & 80.0 & 100.0 & 80.0 \\
\midrule
\multirow{3}{*}{bike-sharing-demand}
 & Helpful selection\% & 0.0 $\pm$ 0.0 & 0.0 $\pm$ 0.0 & 13.3 $\pm$ 26.7 \\
 & Adversarial selection\% & 66.7 $\pm$ 21.1 & 73.3 $\pm$ 13.3 & 100.0 $\pm$ 0.0 \\
 & Valid submission\% & 40.0 & 60.0 & 20.0 \\
\midrule
\multirow{3}{*}{bnp-paribas-cardif-claims-management}
 & Helpful selection\% & 0.0 $\pm$ 0.0 & 0.0 $\pm$ 0.0 & 0.0 $\pm$ 0.0 \\
 & Adversarial selection\% & 100.0 $\pm$ 0.0 & 100.0 $\pm$ 0.0 & 100.0 $\pm$ 0.0 \\
 & Valid submission\% & 20.0 & 20.0 & 20.0 \\
\midrule
\multirow{3}{*}{otto-group-product-classification-challenge}
 & Helpful selection\% & 0.0 $\pm$ 0.0 & 0.0 $\pm$ 0.0 & 0.0 $\pm$ 0.0 \\
 & Adversarial selection\% & 100.0 $\pm$ 0.0 & 100.0 $\pm$ 0.0 & 100.0 $\pm$ 0.0 \\
 & Valid submission\% & 0.0 & 0.0 & 0.0 \\
\midrule
\multirow{3}{*}{rossmann-store-sales}
 & Helpful selection\% & 0.0 $\pm$ 0.0 & 0.0 $\pm$ 0.0 & 0.0 $\pm$ 0.0 \\
 & Adversarial selection\% & 100.0 $\pm$ 0.0 & 100.0 $\pm$ 0.0 & 100.0 $\pm$ 0.0 \\
 & Valid submission\% & 0.0 & 0.0 & 0.0 \\
\bottomrule
\end{tabular}
}
\label{tab:task6_gpt-4o_temp0.7}
\end{table*}

\begin{table*}[!htb]
\centering
\caption{Selection rates of adversarial notebooks by GPT-o4-mini across different datasets with varying helpful to adversarial notebook ratios. The table shows consistent preference for adversarial notebooks (nearly 100\% selection rate) regardless of dataset or notebook ratio, with minimal selection of helpful notebooks.}
\scalebox{0.9}{
\begin{tabular}{ccccc}
\toprule
\textbf{Dataset} & \textbf{Metric} & \multicolumn{3}{c}{\textbf{Helpful : Adversarial}} \\ \cline{3-5}
 &  & \textbf{1:3} & \textbf{2:3} & \textbf{3:1} \\
\midrule
\multirow{3}{*}{allstate-claims-severity}
 & Helpful selection\% & 0.0 $\pm$ 0.0 & 0.0 $\pm$ 0.0 & 0.0 $\pm$ 0.0 \\
 & Adversarial selection\% & 100.0 $\pm$ 0.0 & 100.0 $\pm$ 0.0 & 100.0 $\pm$ 0.0 \\
 & Valid submission\% & 80.0 & 40.0 & 100.0 \\
\midrule
\multirow{3}{*}{bike-sharing-demand}
 & Helpful selection\% & 0.0 $\pm$ 0.0 & 0.0 $\pm$ 0.0 & 0.0 $\pm$ 0.0 \\
 & Adversarial selection\% & 46.7 $\pm$ 26.7 & 73.3 $\pm$ 32.7 & 100.0 $\pm$ 0.0 \\
 & Valid submission\% & 60.0 & 80.0 & 40.0 \\
\midrule
\multirow{3}{*}{bnp-paribas-cardif-claims-management}
 & Helpful selection\% & 0.0 $\pm$ 0.0 & 0.0 $\pm$ 0.0 & 0.0 $\pm$ 0.0 \\
 & Adversarial selection\% & 60.0 $\pm$ 32.7 & 73.3 $\pm$ 32.7 & 100.0 $\pm$ 0.0 \\
 & Valid submission\% & 100.0 & 100.0 & 80.0 \\
\midrule
\multirow{3}{*}{otto-group-product-classification-challenge}
 & Helpful selection\% & 0.0 $\pm$ 0.0 & 0.0 $\pm$ 0.0 & 0.0 $\pm$ 0.0 \\
 & Adversarial selection\% & 100.0 $\pm$ 0.0 & 100.0 $\pm$ 0.0 & 100.0 $\pm$ 0.0 \\
 & Valid submission\% & 20.0 & 20.0 & 20.0 \\
\midrule
\multirow{3}{*}{rossmann-store-sales}
 & Helpful selection\% & 0.0 $\pm$ 0.0 & 0.0 $\pm$ 0.0 & 0.0 $\pm$ 0.0 \\
 & Adversarial selection\% & 100.0 $\pm$ 0.0 & 100.0 $\pm$ 0.0 & 100.0 $\pm$ 0.0 \\
 & Valid submission\% & 0.0 & 20.0 & 20.0 \\
\bottomrule
\end{tabular}
}
\label{tab:task6_o4-mini_temp0.7}
\end{table*}

\begin{table*}[htbp]
\centering
\caption{Kaggle dataset results with \texttt{temperature = $0.7$}, GPT-4o-mini and \texttt{high\_1}, \texttt{high\_2}, \texttt{high\_3} (Task 1).}
\scalebox{0.9}{
\begin{tabular}{ccccc}
\toprule
\textbf{Dataset} & \textbf{Metric} &   \multicolumn{3}{c}{\textbf{No. of Helpful Documents}} \\ \cline{3-5}
 & & \textbf{1} & \textbf{2} & \textbf{3} \\
\midrule
\multirow{2}{*}{allstate-claims-severity}
 & Helpful selection\% &  100.0 $\pm$ 0.0 & 80.0 $\pm$ 24.5 & 86.7 $\pm$ 16.3 \\
 & Adversarial selection\% &   &  &  \\
\midrule
\multirow{2}{*}{bike-sharing-demand}
 & Helpful selection\% &  100.0 $\pm$ 0.0 & 80.0 $\pm$ 24.5 & 73.3 $\pm$ 13.3 \\
 & Adversarial selection\% &   &  &  \\
\midrule
\multirow{2}{*}{bnp-paribas-cardif-claims-management}
 & Helpful selection\% &  0.0 $\pm$ 0.0 & 0.0 $\pm$ 0.0 & 0.0 $\pm$ 0.0 \\
 & Adversarial selection\% &   &  &  \\
\midrule
\multirow{2}{*}{otto-group-product-classification-challenge}
 & Helpful selection\% &  60.0 $\pm$ 49.0 & 50.0 $\pm$ 0.0 & 66.7 $\pm$ 21.1 \\
 & Adversarial selection\% &   &  &  \\
\midrule
\multirow{2}{*}{rossmann-store-sales}
 & Helpful selection\% &  20.0 $\pm$ 40.0 & 100.0 $\pm$ 0.0 & 93.3 $\pm$ 13.3 \\
 & Adversarial selection\% &   &  &  \\
\bottomrule
\end{tabular}
}
\label{tab:task1_gpt-4o-mini_temp0.7}
\end{table*}

\begin{table*}[htbp]
\centering
\caption{Kaggle dataset results with \texttt{temperature = $0.7$}, GPT-4o and \texttt{high\_1}, \texttt{high\_2}, \texttt{high\_3} (Task 1).}
\scalebox{0.9}{
\begin{tabular}{ccccc}
\toprule
\textbf{Dataset} & \textbf{Metric} &   \multicolumn{3}{c}{\textbf{No. of Helpful Documents}} \\ \cline{3-5}
 & & \textbf{1} & \textbf{2} & \textbf{3} \\
\midrule
\multirow{2}{*}{allstate-claims-severity}
 & Helpful selection\% &  100.0 $\pm$ 0.0 & 60.0 $\pm$ 20.0 & 66.7 $\pm$ 21.1 \\
 & Adversarial selection\% &   &  &  \\
\midrule
\multirow{2}{*}{bike-sharing-demand}
 & Helpful selection\% &  100.0 $\pm$ 0.0 & 70.0 $\pm$ 24.5 & 33.3 $\pm$ 0.0 \\
 & Adversarial selection\% &   &  &  \\
\midrule
\multirow{2}{*}{bnp-paribas-cardif-claims-management}
 & Helpful selection\% &  0.0 $\pm$ 0.0 & 0.0 $\pm$ 0.0 & 0.0 $\pm$ 0.0 \\
 & Adversarial selection\% &   &  &  \\
\midrule
\multirow{2}{*}{otto-group-product-classification-challenge}
 & Helpful selection\% &  100.0 $\pm$ 0.0 & 60.0 $\pm$ 20.0 & 66.7 $\pm$ 21.1 \\
 & Adversarial selection\% &   &  &  \\
\midrule
\multirow{2}{*}{rossmann-store-sales}
 & Helpful selection\% &  100.0 $\pm$ 0.0 & 50.0 $\pm$ 0.0 & 66.7 $\pm$ 21.1 \\
 & Adversarial selection\% &   &  &  \\
\bottomrule
\end{tabular}
}
\label{tab:task1_gpt-4o_temp0.7}
\end{table*}

\begin{table*}[htbp]
\centering
\caption{Kaggle dataset results with GPT-o4-mini and \texttt{high\_1}, \texttt{high\_2}, \texttt{high\_3} (Task 1).}
\scalebox{0.9}{
\begin{tabular}{ccccc}
\toprule
\textbf{Dataset} & \textbf{Metric} &   \multicolumn{3}{c}{\textbf{No. of Helpful Documents}} \\ \cline{3-5}
 & & \textbf{1} & \textbf{2} & \textbf{3} \\
\midrule
\multirow{2}{*}{allstate-claims-severity}
 & Helpful selection\% &  100.0 $\pm$ 0.0 & 50.0 $\pm$ 0.0 & 33.3 $\pm$ 0.0 \\
 & Adversarial selection\% &   &  &  \\
\midrule
\multirow{2}{*}{bike-sharing-demand}
 & Helpful selection\% &  80.0 $\pm$ 40.0 & 50.0 $\pm$ 0.0 & 40.0 $\pm$ 13.3 \\
 & Adversarial selection\% &   &  &  \\
\midrule
\multirow{2}{*}{bnp-paribas-cardif-claims-management}
 & Helpful selection\% &  20.0 $\pm$ 40.0 & 40.0 $\pm$ 20.0 & 66.7 $\pm$ 21.1 \\
 & Adversarial selection\% &   &  &  \\
\midrule
\multirow{2}{*}{otto-group-product-classification-challenge}
 & Helpful selection\% &  0.0 $\pm$ 0.0 & 20.0 $\pm$ 24.5 & 6.7 $\pm$ 13.3 \\
 & Adversarial selection\% &   &  &  \\
\midrule
\multirow{2}{*}{rossmann-store-sales}
 & Helpful selection\% &  0.0 $\pm$ 0.0 & 0.0 $\pm$ 0.0 & 20.0 $\pm$ 16.3 \\
 & Adversarial selection\% &   &  &  \\
\bottomrule
\end{tabular}
}
\label{tab:task1_o4-mini_temp0.7}
\end{table*}

\begin{table*}[htbp]
\centering
\caption{Kaggle dataset results with \texttt{temperature = $0.7$}, GPT-4o-mini and \texttt{high\_1\_low\_3}, \texttt{high\_2\_low\_3}, \texttt{high\_3\_low\_1} (Task 2).}
\scalebox{0.9}{
\begin{tabular}{ccccc}
\toprule
\textbf{Dataset} & \textbf{Metric} &   \multicolumn{3}{c}{\textbf{Helpful : Low}} \\ \cline{3-5}
 & & \textbf{1:3} & \textbf{2:3} & \textbf{3:1} \\
\midrule
\multirow{2}{*}{allstate-claims-severity}
 & Helpful selection\% &  100.0 $\pm$ 0.0 & 70.0 $\pm$ 24.5 & 80.0 $\pm$ 16.3 \\
 & Adversarial selection\% &   &  &  \\
\midrule
\multirow{2}{*}{bike-sharing-demand}
 & Helpful selection\% &  100.0 $\pm$ 0.0 & 80.0 $\pm$ 24.5 & 66.7 $\pm$ 0.0 \\
 & Adversarial selection\% &   &  &  \\
\midrule
\multirow{2}{*}{bnp-paribas-cardif-claims-management}
 & Helpful selection\% &  0.0 $\pm$ 0.0 & 0.0 $\pm$ 0.0 & 0.0 $\pm$ 0.0 \\
 & Adversarial selection\% &   &  &  \\
\midrule
\multirow{2}{*}{otto-group-product-classification-challenge}
 & Helpful selection\% &  0.0 $\pm$ 0.0 & 50.0 $\pm$ 44.7 & 26.7 $\pm$ 32.7 \\
 & Adversarial selection\% &   &  &  \\
\midrule
\multirow{2}{*}{rossmann-store-sales}
 & Helpful selection\% &  100.0 $\pm$ 0.0 & 60.0 $\pm$ 49.0 & 93.3 $\pm$ 13.3 \\
 & Adversarial selection\% &   &  &  \\
\bottomrule
\end{tabular}
}
\label{tab:task2_gpt-4o-mini_temp0.7}
\end{table*}

\vspace{1cm}

\begin{table*}[htbp]
\centering
\caption{Kaggle dataset results with \texttt{temperature = $0.7$}, GPT-4o and \texttt{high\_1\_low\_3}, \texttt{high\_2\_low\_3}, \texttt{high\_3\_low\_1} (Task 2).}
\scalebox{0.9}{
\begin{tabular}{ccccc}
\toprule
\textbf{Dataset} & \textbf{Metric} &   \multicolumn{3}{c}{\textbf{Helpful : Low}} \\ \cline{3-5}
 & & \textbf{1:3} & \textbf{2:3} & \textbf{3:1} \\
\midrule
\multirow{2}{*}{allstate-claims-severity}
 & Helpful selection\% &  80.0 $\pm$ 40.0 & 70.0 $\pm$ 24.5 & 66.7 $\pm$ 0.0 \\
 & Adversarial selection\% &   &  &  \\
\midrule
\multirow{2}{*}{bike-sharing-demand}
 & Helpful selection\% &  20.0 $\pm$ 40.0 & 10.0 $\pm$ 20.0 & 60.0 $\pm$ 13.3 \\
 & Adversarial selection\% &   &  &  \\
\midrule
\multirow{2}{*}{bnp-paribas-cardif-claims-management}
 & Helpful selection\% &  0.0 $\pm$ 0.0 & 0.0 $\pm$ 0.0 & 0.0 $\pm$ 0.0 \\
 & Adversarial selection\% &   &  &  \\
\midrule
\multirow{2}{*}{otto-group-product-classification-challenge}
 & Helpful selection\% &  20.0 $\pm$ 40.0 & 10.0 $\pm$ 20.0 & 53.3 $\pm$ 16.3 \\
 & Adversarial selection\% &   &  &  \\
\midrule
\multirow{2}{*}{rossmann-store-sales}
 & Helpful selection\% &  20.0 $\pm$ 40.0 & 20.0 $\pm$ 24.5 & 46.7 $\pm$ 16.3 \\
 & Adversarial selection\% &   &  &  \\
\bottomrule
\end{tabular}
}
\label{tab:task2_gpt-4o_temp0.7}
\end{table*}

\begin{table*}[htbp]
\centering
\caption{Kaggle dataset results with GPT-o4-mini and \texttt{high\_1\_low\_3}, \texttt{high\_2\_low\_3}, \texttt{high\_3\_low\_1} (Task 2).}
\scalebox{0.9}{
\begin{tabular}{ccccc}
\toprule
\textbf{Dataset} & \textbf{Metric} &   \multicolumn{3}{c}{\textbf{Helpful : Low}} \\ \cline{3-5}
 & & \textbf{1:3} & \textbf{2:3} & \textbf{3:1} \\
\midrule
\multirow{2}{*}{allstate-claims-severity}
 & Helpful selection\% &  100.0 $\pm$ 0.0 & 50.0 $\pm$ 0.0 & 33.3 $\pm$ 0.0 \\
 & Adversarial selection\% &   &  &  \\
\midrule
\multirow{2}{*}{bike-sharing-demand}
 & Helpful selection\% &  80.0 $\pm$ 40.0 & 60.0 $\pm$ 20.0 & 33.3 $\pm$ 0.0 \\
 & Adversarial selection\% &   &  &  \\
\midrule
\multirow{2}{*}{bnp-paribas-cardif-claims-management}
 & Helpful selection\% &  0.0 $\pm$ 0.0 & 50.0 $\pm$ 31.6 & 40.0 $\pm$ 13.3 \\
 & Adversarial selection\% &   &  &  \\
\midrule
\multirow{2}{*}{otto-group-product-classification-challenge}
 & Helpful selection\% &  0.0 $\pm$ 0.0 & 10.0 $\pm$ 20.0 & 0.0 $\pm$ 0.0 \\
 & Adversarial selection\% &   &  &  \\
\midrule
\multirow{2}{*}{rossmann-store-sales}
 & Helpful selection\% &  0.0 $\pm$ 0.0 & 10.0 $\pm$ 20.0 & 0.0 $\pm$ 0.0 \\
 & Adversarial selection\% &   &  &  \\
\bottomrule
\end{tabular}
}
\label{tab:task2_o4-mini_temp0.7}
\end{table*}

\begin{table*}[htbp]
\centering
\caption{Kaggle dataset results with \texttt{temperature = $0.7$}, GPT-4o-mini and \texttt{high\_1\_adv\_3\_neu\_prompt}, \texttt{high\_2\_adv\_3\_neu\_prompt}, \texttt{high\_3\_adv\_1\_neu\_prompt} (Task 3).}
\scalebox{0.9}{
\begin{tabular}{ccccc}
\toprule
\textbf{Dataset} & \textbf{Metric} &   \multicolumn{3}{c}{\textbf{Helpful : Adversarial}} \\ \cline{3-5}
 & & \textbf{1:3} & \textbf{2:3} & \textbf{3:1} \\
\midrule
\multirow{2}{*}{allstate-claims-severity}
 & Helpful selection\% &  100.0 $\pm$ 0.0 & 50.0 $\pm$ 0.0 & 86.7 $\pm$ 16.3 \\
 & Adversarial selection\% &  80.0 $\pm$ 26.7 & 53.3 $\pm$ 16.3 & 80.0 $\pm$ 40.0 \\
\midrule
\multirow{2}{*}{bike-sharing-demand}
 & Helpful selection\% &  100.0 $\pm$ 0.0 & 100.0 $\pm$ 0.0 & 66.7 $\pm$ 0.0 \\
 & Adversarial selection\% &  60.0 $\pm$ 13.3 & 53.3 $\pm$ 26.7 & 100.0 $\pm$ 0.0 \\
\midrule
\multirow{2}{*}{bnp-paribas-cardif-claims-management}
 & Helpful selection\% &  0.0 $\pm$ 0.0 & 0.0 $\pm$ 0.0 & 0.0 $\pm$ 0.0 \\
 & Adversarial selection\% &  93.3 $\pm$ 13.3 & 86.7 $\pm$ 26.7 & 100.0 $\pm$ 0.0 \\
\midrule
\multirow{2}{*}{otto-group-product-classification-challenge}
 & Helpful selection\% &  0.0 $\pm$ 0.0 & 40.0 $\pm$ 20.0 & 93.3 $\pm$ 13.3 \\
 & Adversarial selection\% &  93.3 $\pm$ 13.3 & 13.3 $\pm$ 16.3 & 60.0 $\pm$ 49.0 \\
\midrule
\multirow{2}{*}{rossmann-store-sales}
 & Helpful selection\% &  60.0 $\pm$ 49.0 & 0.0 $\pm$ 0.0 & 40.0 $\pm$ 24.9 \\
 & Adversarial selection\% &  73.3 $\pm$ 38.9 & 93.3 $\pm$ 13.3 & 80.0 $\pm$ 40.0 \\
\bottomrule
\end{tabular}
}
\label{tab:task3_gpt-4o-mini_temp0.7}
\end{table*}

\begin{table*}[htbp]
\centering
\caption{Kaggle dataset results with \texttt{temperature = $0.7$}, GPT-4o and \texttt{high\_1\_adv\_3\_neu\_prompt}, \texttt{high\_2\_adv\_3\_neu\_prompt}, \texttt{high\_3\_adv\_1\_neu\_prompt} (Task 3).}
\scalebox{0.9}{
\begin{tabular}{ccccc}
\toprule
\textbf{Dataset} & \textbf{Metric} &   \multicolumn{3}{c}{\textbf{Helpful : Adversarial}} \\ \cline{3-5}
 & & \textbf{1:3} & \textbf{2:3} & \textbf{3:1} \\
\midrule
\multirow{2}{*}{allstate-claims-severity}
 & Helpful selection\% &  0.0 $\pm$ 0.0 & 20.0 $\pm$ 24.5 & 73.3 $\pm$ 24.9 \\
 & Adversarial selection\% &  100.0 $\pm$ 0.0 & 100.0 $\pm$ 0.0 & 20.0 $\pm$ 40.0 \\
\midrule
\multirow{2}{*}{bike-sharing-demand}
 & Helpful selection\% &  80.0 $\pm$ 40.0 & 50.0 $\pm$ 31.6 & 60.0 $\pm$ 24.9 \\
 & Adversarial selection\% &  73.3 $\pm$ 32.7 & 60.0 $\pm$ 32.7 & 100.0 $\pm$ 0.0 \\
\midrule
\multirow{2}{*}{bnp-paribas-cardif-claims-management}
 & Helpful selection\% &  0.0 $\pm$ 0.0 & 0.0 $\pm$ 0.0 & 0.0 $\pm$ 0.0 \\
 & Adversarial selection\% &  93.3 $\pm$ 13.3 & 100.0 $\pm$ 0.0 & 100.0 $\pm$ 0.0 \\
\midrule
\multirow{2}{*}{otto-group-product-classification-challenge}
 & Helpful selection\% &  0.0 $\pm$ 0.0 & 0.0 $\pm$ 0.0 & 13.3 $\pm$ 16.3 \\
 & Adversarial selection\% &  93.3 $\pm$ 13.3 & 100.0 $\pm$ 0.0 & 100.0 $\pm$ 0.0 \\
\midrule
\multirow{2}{*}{rossmann-store-sales}
 & Helpful selection\% &  0.0 $\pm$ 0.0 & 0.0 $\pm$ 0.0 & 0.0 $\pm$ 0.0 \\
 & Adversarial selection\% &  100.0 $\pm$ 0.0 & 73.3 $\pm$ 24.9 & 100.0 $\pm$ 0.0 \\
\bottomrule
\end{tabular}
}
\label{tab:task3_gpt-4o_temp0.7}
\end{table*}

\begin{table*}[htbp]
\centering
\caption{Kaggle dataset results with GPT-o4-mini and \texttt{high\_1\_adv\_3\_neu\_prompt}, \texttt{high\_2\_adv\_3\_neu\_prompt}, \texttt{high\_3\_adv\_1\_neu\_prompt} (Task 3).}
\scalebox{0.9}{
\begin{tabular}{ccccc}
\toprule
\textbf{Dataset} & \textbf{Metric} &   \multicolumn{3}{c}{\textbf{Helpful : Adversarial}} \\ \cline{3-5}
 & & \textbf{1:3} & \textbf{2:3} & \textbf{3:1} \\
\midrule
\multirow{2}{*}{allstate-claims-severity}
 & Helpful selection\% &  100.0 $\pm$ 0.0 & 50.0 $\pm$ 0.0 & 33.3 $\pm$ 0.0 \\
 & Adversarial selection\% &  0.0 $\pm$ 0.0 & 0.0 $\pm$ 0.0 & 0.0 $\pm$ 0.0 \\
\midrule
\multirow{2}{*}{bike-sharing-demand}
 & Helpful selection\% &  80.0 $\pm$ 40.0 & 50.0 $\pm$ 0.0 & 33.3 $\pm$ 0.0 \\
 & Adversarial selection\% &  6.7 $\pm$ 13.3 & 0.0 $\pm$ 0.0 & 0.0 $\pm$ 0.0 \\
\midrule
\multirow{2}{*}{bnp-paribas-cardif-claims-management}
 & Helpful selection\% &  0.0 $\pm$ 0.0 & 10.0 $\pm$ 20.0 & 20.0 $\pm$ 26.7 \\
 & Adversarial selection\% &  60.0 $\pm$ 38.9 & 100.0 $\pm$ 0.0 & 100.0 $\pm$ 0.0 \\
\midrule
\multirow{2}{*}{otto-group-product-classification-challenge}
 & Helpful selection\% &  0.0 $\pm$ 0.0 & 10.0 $\pm$ 20.0 & 0.0 $\pm$ 0.0 \\
 & Adversarial selection\% &  20.0 $\pm$ 40.0 & 6.7 $\pm$ 13.3 & 80.0 $\pm$ 40.0 \\
\midrule
\multirow{2}{*}{rossmann-store-sales}
 & Helpful selection\% &  0.0 $\pm$ 0.0 & 20.0 $\pm$ 24.5 & 6.7 $\pm$ 13.3 \\
 & Adversarial selection\% &  40.0 $\pm$ 49.0 & 40.0 $\pm$ 49.0 & 60.0 $\pm$ 49.0 \\
\bottomrule
\end{tabular}
}
\label{tab:task3_o4-mini_temp0.7}
\end{table*}

\begin{table*}[htbp]
\centering
\caption{Kaggle dataset results with \texttt{temperature = $0.7$}, GPT-4o-mini and \texttt{high\_1\_adv\_3\_helpful\_prompt}, \texttt{high\_2\_adv\_3\_helpful\_prompt}, \texttt{high\_3\_adv\_1\_helpful\_prompt} (Task 4).}
\scalebox{0.9}{
\begin{tabular}{ccccc}
\toprule
\textbf{Dataset} & \textbf{Metric} &   \multicolumn{3}{c}{\textbf{Helpful : Adversarial}} \\ \cline{3-5}
 & & \textbf{1:3} & \textbf{2:3} & \textbf{3:1} \\
\midrule
\multirow{2}{*}{allstate-claims-severity}
 & Helpful selection\% &  100.0 $\pm$ 0.0 & 100.0 $\pm$ 0.0 & 100.0 $\pm$ 0.0 \\
 & Adversarial selection\% &  0.0 $\pm$ 0.0 & 33.3 $\pm$ 0.0 & 0.0 $\pm$ 0.0 \\
\midrule
\multirow{2}{*}{bike-sharing-demand}
 & Helpful selection\% &  100.0 $\pm$ 0.0 & 100.0 $\pm$ 0.0 & 80.0 $\pm$ 16.3 \\
 & Adversarial selection\% &  13.3 $\pm$ 26.7 & 33.3 $\pm$ 0.0 & 40.0 $\pm$ 49.0 \\
\midrule
\multirow{2}{*}{bnp-paribas-cardif-claims-management}
 & Helpful selection\% &  0.0 $\pm$ 0.0 & 0.0 $\pm$ 0.0 & 0.0 $\pm$ 0.0 \\
 & Adversarial selection\% &  93.3 $\pm$ 13.3 & 86.7 $\pm$ 16.3 & 100.0 $\pm$ 0.0 \\
\midrule
\multirow{2}{*}{otto-group-product-classification-challenge}
 & Helpful selection\% &  0.0 $\pm$ 0.0 & 50.0 $\pm$ 31.6 & 66.7 $\pm$ 0.0 \\
 & Adversarial selection\% &  46.7 $\pm$ 45.2 & 26.7 $\pm$ 13.3 & 100.0 $\pm$ 0.0 \\
\midrule
\multirow{2}{*}{rossmann-store-sales}
 & Helpful selection\% &  100.0 $\pm$ 0.0 & 20.0 $\pm$ 40.0 & 60.0 $\pm$ 32.7 \\
 & Adversarial selection\% &  60.0 $\pm$ 32.7 & 80.0 $\pm$ 26.7 & 80.0 $\pm$ 40.0 \\
\bottomrule
\end{tabular}
}
\label{tab:task4_gpt-4o-mini_temp0.7}
\end{table*}

\begin{table*}[htbp]
\centering
\caption{Kaggle dataset results with \texttt{temperature = $0.7$}, GPT-4o and \texttt{high\_1\_adv\_3\_helpful\_prompt}, \texttt{high\_2\_adv\_3\_helpful\_prompt}, \texttt{high\_3\_adv\_1\_helpful\_prompt} (Task 4).}
\scalebox{0.9}{
\begin{tabular}{ccccc}
\toprule
\textbf{Dataset} & \textbf{Metric} &   \multicolumn{3}{c}{\textbf{Helpful : Adversarial}} \\ \cline{3-5}
 & & \textbf{1:3} & \textbf{2:3} & \textbf{3:1} \\
\midrule
\multirow{2}{*}{allstate-claims-severity}
 & Helpful selection\% &  100.0 $\pm$ 0.0 & 100.0 $\pm$ 0.0 & 53.3 $\pm$ 16.3 \\
 & Adversarial selection\% &  0.0 $\pm$ 0.0 & 0.0 $\pm$ 0.0 & 0.0 $\pm$ 0.0 \\
\midrule
\multirow{2}{*}{bike-sharing-demand}
 & Helpful selection\% &  100.0 $\pm$ 0.0 & 80.0 $\pm$ 24.5 & 86.7 $\pm$ 16.3 \\
 & Adversarial selection\% &  0.0 $\pm$ 0.0 & 0.0 $\pm$ 0.0 & 0.0 $\pm$ 0.0 \\
\midrule
\multirow{2}{*}{bnp-paribas-cardif-claims-management}
 & Helpful selection\% &  0.0 $\pm$ 0.0 & 0.0 $\pm$ 0.0 & 0.0 $\pm$ 0.0 \\
 & Adversarial selection\% &  0.0 $\pm$ 0.0 & 0.0 $\pm$ 0.0 & 40.0 $\pm$ 49.0 \\
\midrule
\multirow{2}{*}{otto-group-product-classification-challenge}
 & Helpful selection\% &  100.0 $\pm$ 0.0 & 30.0 $\pm$ 24.5 & 73.3 $\pm$ 13.3 \\
 & Adversarial selection\% &  0.0 $\pm$ 0.0 & 33.3 $\pm$ 36.5 & 80.0 $\pm$ 40.0 \\
\midrule
\multirow{2}{*}{rossmann-store-sales}
 & Helpful selection\% &  20.0 $\pm$ 40.0 & 40.0 $\pm$ 37.4 & 26.7 $\pm$ 24.9 \\
 & Adversarial selection\% &  66.7 $\pm$ 42.2 & 40.0 $\pm$ 49.0 & 100.0 $\pm$ 0.0 \\
\bottomrule
\end{tabular}
}
\label{tab:task4_gpt-4o_temp0.7}
\end{table*}

\begin{table*}[htbp]
\centering
\caption{Kaggle dataset results with  GPT-o4-mini and \texttt{high\_1\_adv\_3\_helpful\_prompt}, \texttt{high\_2\_adv\_3\_helpful\_prompt}, \texttt{high\_3\_adv\_1\_helpful\_prompt} (Task 4).}
\scalebox{0.9}{
\begin{tabular}{ccccc}
\toprule
\textbf{Dataset} & \textbf{Metric} &   \multicolumn{3}{c}{\textbf{Helpful : Adversarial}} \\ \cline{3-5}
 & & \textbf{1:3} & \textbf{2:3} & \textbf{3:1} \\
\midrule
\multirow{2}{*}{allstate-claims-severity}
 & Helpful selection\% &  100.0 $\pm$ 0.0 & 70.0 $\pm$ 24.5 & 40.0 $\pm$ 13.3 \\
 & Adversarial selection\% &  0.0 $\pm$ 0.0 & 0.0 $\pm$ 0.0 & 0.0 $\pm$ 0.0 \\
\midrule
\multirow{2}{*}{bike-sharing-demand}
 & Helpful selection\% &  100.0 $\pm$ 0.0 & 80.0 $\pm$ 24.5 & 40.0 $\pm$ 13.3 \\
 & Adversarial selection\% &  0.0 $\pm$ 0.0 & 0.0 $\pm$ 0.0 & 0.0 $\pm$ 0.0 \\
\midrule
\multirow{2}{*}{bnp-paribas-cardif-claims-management}
 & Helpful selection\% &  60.0 $\pm$ 49.0 & 90.0 $\pm$ 20.0 & 33.3 $\pm$ 29.8 \\
 & Adversarial selection\% &  0.0 $\pm$ 0.0 & 0.0 $\pm$ 0.0 & 60.0 $\pm$ 49.0 \\
\midrule
\multirow{2}{*}{otto-group-product-classification-challenge}
 & Helpful selection\% &  40.0 $\pm$ 49.0 & 40.0 $\pm$ 37.4 & 26.7 $\pm$ 24.9 \\
 & Adversarial selection\% &  0.0 $\pm$ 0.0 & 0.0 $\pm$ 0.0 & 20.0 $\pm$ 40.0 \\
\midrule
\multirow{2}{*}{rossmann-store-sales}
 & Helpful selection\% &  0.0 $\pm$ 0.0 & 40.0 $\pm$ 37.4 & 40.0 $\pm$ 24.9 \\
 & Adversarial selection\% &  0.0 $\pm$ 0.0 & 0.0 $\pm$ 0.0 & 0.0 $\pm$ 0.0 \\
\bottomrule
\end{tabular}
}
\label{tab:task4_o4-mini_temp0.7}
\end{table*}

\begin{table*}[htbp]
\centering
\caption{Kaggle dataset results with \texttt{temperature = $0.7$}, GPT-4o-mini and \texttt{high\_1\_adv\_3\_specific\_helpful\_prompt}, \texttt{high\_2\_adv\_3\_specific\_helpful\_prompt}, \texttt{high\_3\_adv\_1\_specific\_helpful\_prompt} (Task 5).}
\scalebox{0.9}{
\begin{tabular}{ccccc}
\toprule
\textbf{Dataset} & \textbf{Metric} &   \multicolumn{3}{c}{\textbf{Helpful : Adversarial}} \\ \cline{3-5}
 & & \textbf{1:3} & \textbf{2:3} & \textbf{3:1} \\
\midrule
\multirow{2}{*}{allstate-claims-severity}
 & Helpful selection\% &  100.0 $\pm$ 0.0 & 100.0 $\pm$ 0.0 & 93.3 $\pm$ 13.3 \\
 & Adversarial selection\% &  0.0 $\pm$ 0.0 & 0.0 $\pm$ 0.0 & 0.0 $\pm$ 0.0 \\
\midrule
\multirow{2}{*}{bike-sharing-demand}
 & Helpful selection\% &  100.0 $\pm$ 0.0 & 100.0 $\pm$ 0.0 & 100.0 $\pm$ 0.0 \\
 & Adversarial selection\% &  0.0 $\pm$ 0.0 & 33.3 $\pm$ 21.1 & 40.0 $\pm$ 49.0 \\
\midrule
\multirow{2}{*}{bnp-paribas-cardif-claims-management}
 & Helpful selection\% &  0.0 $\pm$ 0.0 & 0.0 $\pm$ 0.0 & 0.0 $\pm$ 0.0 \\
 & Adversarial selection\% &  20.0 $\pm$ 40.0 & 20.0 $\pm$ 26.7 & 0.0 $\pm$ 0.0 \\
\midrule
\multirow{2}{*}{otto-group-product-classification-challenge}
 & Helpful selection\% &  100.0 $\pm$ 0.0 & 100.0 $\pm$ 0.0 & 100.0 $\pm$ 0.0 \\
 & Adversarial selection\% &  0.0 $\pm$ 0.0 & 0.0 $\pm$ 0.0 & 40.0 $\pm$ 49.0 \\
\midrule
\multirow{2}{*}{rossmann-store-sales}
 & Helpful selection\% &  100.0 $\pm$ 0.0 & 100.0 $\pm$ 0.0 & 100.0 $\pm$ 0.0 \\
 & Adversarial selection\% &  13.3 $\pm$ 26.7 & 46.7 $\pm$ 26.7 & 20.0 $\pm$ 40.0 \\
\bottomrule
\end{tabular}
}
\label{tab:task5_gpt-4o-mini_temp0.7}
\end{table*}

\begin{table*}[htbp]
\centering
\caption{Kaggle dataset results with \texttt{temperature = $0.7$}, GPT-4o and \texttt{high\_1\_adv\_3\_specific\_helpful\_prompt}, \texttt{high\_2\_adv\_3\_specific\_helpful\_prompt}, \texttt{high\_3\_adv\_1\_specific\_helpful\_prompt} (Task 5).}
\scalebox{0.9}{
\begin{tabular}{ccccc}
\toprule
\textbf{Dataset} & \textbf{Metric} &   \multicolumn{3}{c}{\textbf{Helpful : Adversarial}} \\ \cline{3-5}
 & & \textbf{1:3} & \textbf{2:3} & \textbf{3:1} \\
\midrule
\multirow{2}{*}{allstate-claims-severity}
 & Helpful selection\% &  100.0 $\pm$ 0.0 & 100.0 $\pm$ 0.0 & 80.0 $\pm$ 26.7 \\
 & Adversarial selection\% &  0.0 $\pm$ 0.0 & 0.0 $\pm$ 0.0 & 0.0 $\pm$ 0.0 \\
\midrule
\multirow{2}{*}{bike-sharing-demand}
 & Helpful selection\% &  100.0 $\pm$ 0.0 & 100.0 $\pm$ 0.0 & 93.3 $\pm$ 13.3 \\
 & Adversarial selection\% &  0.0 $\pm$ 0.0 & 0.0 $\pm$ 0.0 & 20.0 $\pm$ 40.0 \\
\midrule
\multirow{2}{*}{bnp-paribas-cardif-claims-management}
 & Helpful selection\% &  0.0 $\pm$ 0.0 & 0.0 $\pm$ 0.0 & 0.0 $\pm$ 0.0 \\
 & Adversarial selection\% &  0.0 $\pm$ 0.0 & 0.0 $\pm$ 0.0 & 0.0 $\pm$ 0.0 \\
\midrule
\multirow{2}{*}{otto-group-product-classification-challenge}
 & Helpful selection\% &  100.0 $\pm$ 0.0 & 80.0 $\pm$ 24.5 & 60.0 $\pm$ 24.9 \\
 & Adversarial selection\% &  0.0 $\pm$ 0.0 & 0.0 $\pm$ 0.0 & 0.0 $\pm$ 0.0 \\
\midrule
\multirow{2}{*}{rossmann-store-sales}
 & Helpful selection\% &  100.0 $\pm$ 0.0 & 80.0 $\pm$ 24.5 & 66.7 $\pm$ 21.1 \\
 & Adversarial selection\% &  20.0 $\pm$ 40.0 & 0.0 $\pm$ 0.0 & 60.0 $\pm$ 49.0 \\
\bottomrule
\end{tabular}
}
\label{tab:task5_gpt-4o_temp0.7}
\end{table*}

\begin{table*}[htbp]
\centering
\caption{Kaggle dataset results with GPT-o4-mini and \texttt{high\_1\_adv\_3\_specific\_helpful\_prompt}, \texttt{high\_2\_adv\_3\_specific\_helpful\_prompt}, \texttt{high\_3\_adv\_1\_specific\_helpful\_prompt} (Task 5).}
\scalebox{0.9}{
\begin{tabular}{ccccc}
\toprule
\textbf{Dataset} & \textbf{Metric} &   \multicolumn{3}{c}{\textbf{Helpful : Adversarial}} \\ \cline{3-5}
 & & \textbf{1:3} & \textbf{2:3} & \textbf{3:1} \\
\midrule
\multirow{2}{*}{allstate-claims-severity}
 & Helpful selection\% &  100.0 $\pm$ 0.0 & 80.0 $\pm$ 24.5 & 46.7 $\pm$ 16.3 \\
 & Adversarial selection\% &  0.0 $\pm$ 0.0 & 0.0 $\pm$ 0.0 & 0.0 $\pm$ 0.0 \\
\midrule
\multirow{2}{*}{bike-sharing-demand}
 & Helpful selection\% &  100.0 $\pm$ 0.0 & 50.0 $\pm$ 0.0 & 53.3 $\pm$ 26.7 \\
 & Adversarial selection\% &  0.0 $\pm$ 0.0 & 0.0 $\pm$ 0.0 & 0.0 $\pm$ 0.0 \\
\midrule
\multirow{2}{*}{bnp-paribas-cardif-claims-management}
 & Helpful selection\% &  80.0 $\pm$ 40.0 & 90.0 $\pm$ 20.0 & 66.7 $\pm$ 29.8 \\
 & Adversarial selection\% &  0.0 $\pm$ 0.0 & 0.0 $\pm$ 0.0 & 0.0 $\pm$ 0.0 \\
\midrule
\multirow{2}{*}{otto-group-product-classification-challenge}
 & Helpful selection\% &  100.0 $\pm$ 0.0 & 60.0 $\pm$ 37.4 & 26.7 $\pm$ 13.3 \\
 & Adversarial selection\% &  0.0 $\pm$ 0.0 & 0.0 $\pm$ 0.0 & 0.0 $\pm$ 0.0 \\
\midrule
\multirow{2}{*}{rossmann-store-sales}
 & Helpful selection\% &  80.0 $\pm$ 40.0 & 50.0 $\pm$ 31.6 & 33.3 $\pm$ 21.1 \\
 & Adversarial selection\% &  0.0 $\pm$ 0.0 & 0.0 $\pm$ 0.0 & 0.0 $\pm$ 0.0 \\
\bottomrule
\end{tabular}
}
\label{tab:task5_o4-mini_temp0.7}
\end{table*}

\end{document}